\documentclass[11pt]{article}

\usepackage[preprint]{acl}

\usepackage{times}
\usepackage{latexsym}
\usepackage{pifont}
\usepackage{graphicx}
\usepackage{latexsym}
\usepackage{supertabular}
\usepackage{stfloats}
\usepackage{tabularx}
\usepackage{multirow}
\usepackage{makecell}
\usepackage{booktabs}
\usepackage{amsmath}
\usepackage{amsfonts}
\usepackage{amssymb}
\usepackage{enumerate}
\usepackage{algorithm}
\usepackage{algorithmic}
\usepackage{bbm}
\usepackage{pifont}
\usepackage{colortbl}
\usepackage{xspace}
\usepackage{inconsolata}
\usepackage{subfigure}
\usepackage{caption}
\usepackage{subcaption}
\usepackage{enumitem}
\usepackage{tcolorbox}

\usepackage{float}  
\usepackage{subfigure}  
\usepackage{listings}
\usepackage{color}
\usepackage{wrapfig}
\usepackage{url}

\definecolor{dkgreen}{rgb}{0,0.6,0}
\definecolor{gray}{rgb}{0.5,0.5,0.5}
\definecolor{mauve}{rgb}{0.58,0,0.82}
\definecolor{mylcacc}{RGB}{0, 0, 0}
\definecolor{mygray}{RGB}{0, 0, 0}
\definecolor{mylc}{RGB}{0, 0, 0}

\usepackage[T1]{fontenc}

\usepackage[utf8]{inputenc}

\usepackage{microtype}

\usepackage{inconsolata}

\usepackage{graphicx}
\usepackage{amsmath} 

\def\huggingface{\raisebox{-1.5pt}{\includegraphics[height=1.05em]{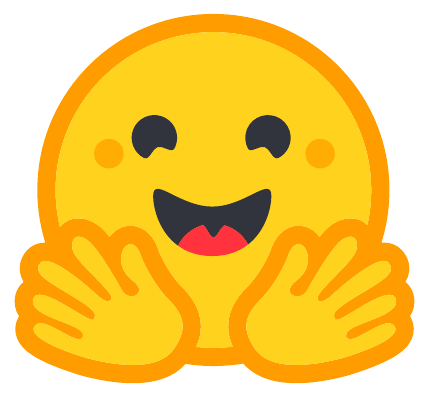}}}
\def\github{\raisebox{-1.5pt}{\includegraphics[height=1.0em]{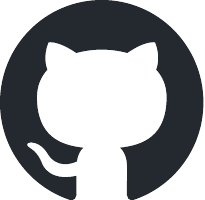}}}

\title{\emph{Think Natively}: Unlocking Multilingual Reasoning with Consistency-Enhanced Reinforcement Learning}

\author{
\github \enspace \href{https://github.com/XZhang00/M-Thinker}{Code} \quad \huggingface \enspace \href{https://huggingface.co/collections/XueZhang-bjtu/m-thinker-68edff19c1d6ae5ca3287a67}{Model} \quad \huggingface \enspace \href{https://huggingface.co/collections/XueZhang-bjtu/m-thinker-data-68ee005490c959d32c70de74}{Data} \\
\textbf{Xue Zhang\textsuperscript{1,2}\thanks{ \ This work was done during the internship at Pattern Recognition Center, WeChat AI, Tencent Inc, China.}}, \textbf{Yunlong Liang\textsuperscript{3}}, 
\textbf{Fandong Meng\textsuperscript{3}}, \textbf{Songming           Zhang\textsuperscript{1,2}},\\
\textbf{Kaiyu Huang\textsuperscript{1,2}}, \textbf{Yufeng Chen\textsuperscript{1,2}\thanks{ \ \ Yufeng Chen is the corresponding author.}},
\textbf{Jinan Xu\textsuperscript{1,2}}, \textbf{Jie Zhou\textsuperscript{3}} \\
\textsuperscript{1}Key Laboratory of Big Data \& Artificial Intelligence in Transportation, Beijing Jiaotong University\\
\textsuperscript{2}School of Computer Science and Technology, Beijing Jiaotong University, Beijing, China \\
\textsuperscript{3}Pattern Recognition Center, WeChat AI, Tencent Inc, China \\
\texttt{\{\text{zhang\_xue},smzhang22,chenyf,jaxu\}@bjtu.edu.cn} \\
}

\begin{document}
\maketitle
\begin{abstract}
Large Reasoning Models (LRMs) have achieved remarkable performance on complex reasoning tasks by adopting the ``think-then-answer'' paradigm, which enhances both accuracy and interpretability.
However, current LRMs exhibit two critical limitations when processing non-English languages: 
(1) They often struggle to maintain input-output language consistency;
(2) They generally perform poorly with wrong reasoning paths and lower answer accuracy compared to English.  
These limitations significantly compromise the interpretability of reasoning processes and degrade the user experience for non-English speakers, hindering the global deployment of LRMs.
To address these limitations, we propose \textbf{M-Thinker}, which is trained by the GRPO algorithm that involves a Language Consistency (LC) reward and a novel Cross-lingual Thinking Alignment (CTA) reward.
Specifically, the LC reward defines a strict constraint on the language consistency between the input, thought, and answer.
Besides, the CTA reward compares the model's non-English reasoning paths with its English reasoning path to transfer its own reasoning capability from English to non-English languages.
Through an iterative RL procedure, our M-Thinker-1.5B/4B/7B models not only achieve nearly 100\% language consistency and superior performance on two multilingual benchmarks (MMATH and PolyMath), but also exhibit excellent generalization on out-of-domain languages.


\end{abstract}

\section{Introduction}









Large reasoning models (LRMs), such as DeepSeek-R1 \cite{deepseekai2025deepseekr1incentivizingreasoningcapability}, OpenAI-o3 \cite{openai2025competitiveprogramminglargereasoning}, and Qwen3 \cite{yang2025qwen3technicalreport}, have achieved impressive performance across a variety of complex reasoning tasks, such as mathematical problem solving, code generation, and logical deduction. 
A key advantage of these models lies in their response pattern: They first generate an explicit chain of reasoning \cite{tam2025languagemattersmultilingualinput} that may include problem decomposition, solution planning, and intermediate verification, and then offer an answer summary.
This ``think-then-answer'' paradigm not only enhances performance but also significantly improves transparency and interpretability of answers \cite{wang2025languagemixingreasoninglanguage}, making the decision-making process more accessible and trustworthy for users.

\begin{figure}[t]
    \centering
    \includegraphics[width=\linewidth]{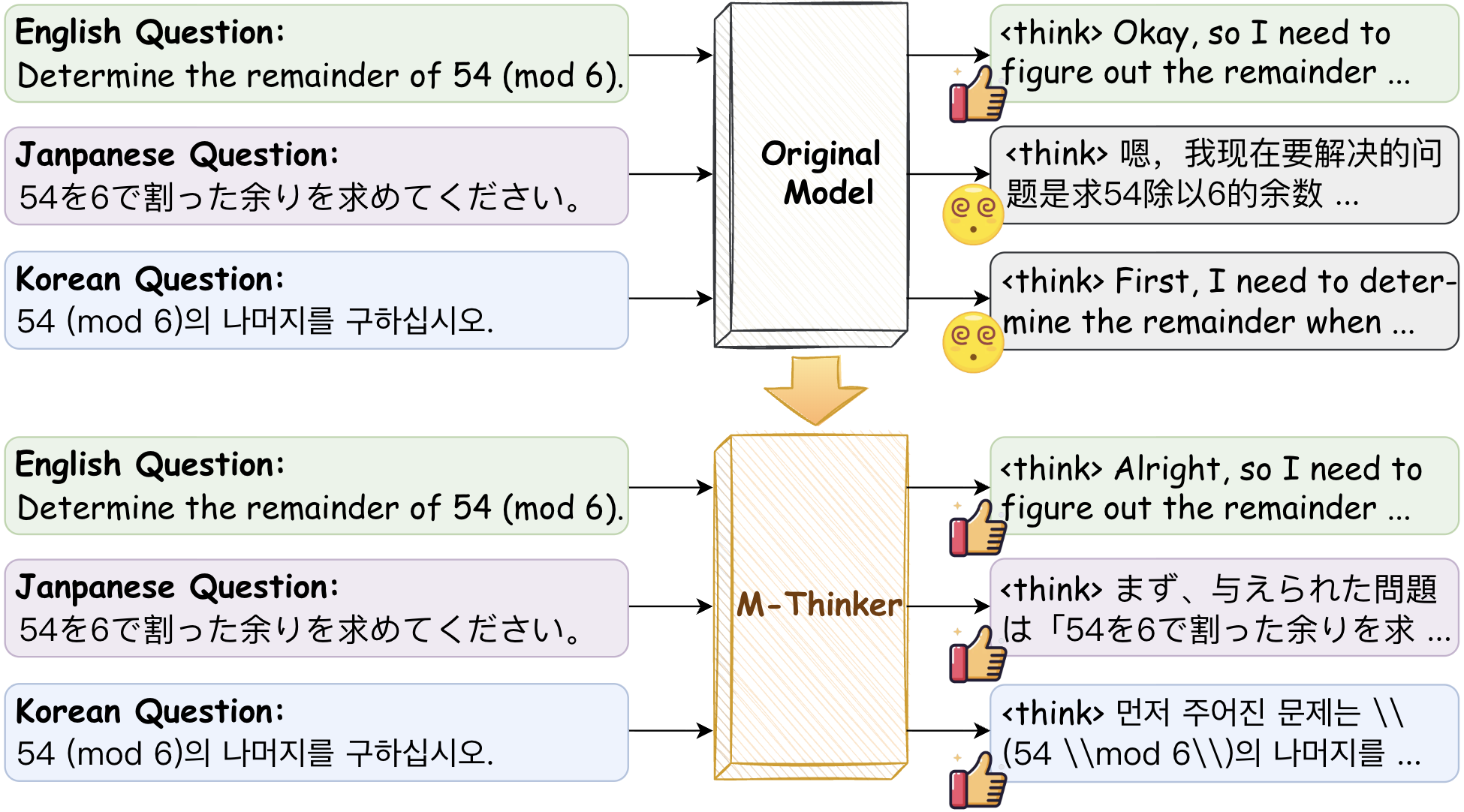}
    \caption{Existing LRMs struggle to maintain input-output language consistency and probably offer us the wrong answer when processing non-English inputs, while our M-Thinker can respond in the input language with the correct answer.}
    \label{fig:mthinker}
\end{figure}

However, current LRMs generally suffer from two major issues under multilingual scenarios.
First, they often suffer from \textbf{input-output language inconsistency} \cite{wang2025polymath, tam2025languagemattersmultilingualinput}, \textit{i.e.}, they frequently default to thinking and answering in English (or other unintended languages) rather than the input language (please refer to Figure \ref{fig:mthinker}).
Second, they present \textbf{inferior performance} for other languages compared to English \cite{luo2025mmathmultilingualbenchmarkmathematical, wang2025polymath}. 
These issues significantly reduce the readability and explainability of the reasoning processes and degrade the user experience of LRMs in multilingual environments.
To mitigate these issues, current solutions include language control instructions \cite{tam2025languagemattersmultilingualinput}, supervised fine-tuning (SFT) with specific language data \cite{luo2025mmathmultilingualbenchmarkmathematical}, and GRPO \cite{shao2024deepseekmathpushinglimitsmathematical} with a soft language reward \cite{park2025crosslingualcollapselanguagecentricfoundation, 2025arXiv250610910M,  hwang2025learngloballyspeaklocally}.
However, these solutions still face notable limitations: Prompt-based methods struggle to enforce output language consistency with the input; SFT generally entails a trade-off between answer accuracy and language consistency; Soft consistency rewards in GRPO can only impose weak constraints on maintaining language consistency.
Therefore, there still remains a clear need for a solution to effectively enhance both language consistency and multilingual reasoning capability of LRMs. 

To this end, we propose M-Thinker, a real multilingual reasoning model trained by the GRPO algorithm that includes a Language Consistency (LC) reward and a novel Cross-lingual Thinking Alignment (CTA) reward.
Specifically, the LC reward strictly constrains the language consistency between the input, thought, and answer, encouraging the model to generate language-consistent responses.
Additionally, given that LRMs often exhibit stronger reasoning proficiency in English compared to other languages \cite{huang2025surveylargelanguagemodels, zhang2025cmalignconsistencybasedmultilingualalignment}, we regard the English reasoning paths of the model itself as the teacher and design the CTA reward for cross-lingual reasoning alignment.
The CTA reward is computed by comparing the model's reasoning paths in English and other languages via LLM-as-a-Judge \cite{gu2025surveyllmasajudge, wang2025deeptransdeepreasoningtranslation}, which encourages the model to transfer its reasoning capability from English to non-English languages.
On this basis, our M-Thinker is trained with a systematic training procedure incorporating cold-start SFT, rejection sampling, and iterative RL training.

Experimental results on two publicly-used multilingual benchmarks (MMATH and PolyMath) show that our M-Thinker-1.5B/4B/7B models not only achieve nearly 100\% language consistency and substantial performance improvement, but also demonstrate remarkable generalization on out-of-domain languages.
In summary, the major contributions of this paper are as follows\footnote{\url{https://github.com/XZhang00/M-Thinker}}:
\begin{itemize}
    \item We propose M-Thinker, which both achieves the input-output language consistency with a Language Consistency reward and enhances the multilingual reasoning performance with a Cross-lingual Thinking Alignment reward.
    \item Experimental results of our M-Thinker-1.5B/4B/7B models on MMATH and PolyMath benchmarks demonstrate superior performance on both language consistency and answer accuracy for multiple languages.
    \item We also conduct an analysis on the generalization of M-Thinker to out-of-domain languages, which reveals that the models typically generalize better to languages within the same or similar language families.
\end{itemize}

\section{Related Work}

The multilingual reasoning capabilities of current LRMs have recently drawn increasing research interest.
\citet{luo2025mmathmultilingualbenchmarkmathematical} point that DeepSeek-R1 exhibits substantial performance disparities across languages and suffers from a critical off-target issue, \textit{i.e.}, generating responses in unintended languages.
\citet{wang2025polymath} also show that reasoning models exhibit lower input-output language consistency, particularly in their thinking processes.
Additionally, when constrained to reason in the same language as the input, the model's performance declines, especially for low-resource languages \cite{tam2025languagemattersmultilingualinput}.
Furthermore, \citet{wang2025languagemixingreasoninglanguage} investigate that the language-mixing phenomenon may affect the final performance, which may hinder the readability and usability of
outputs in multilingual contexts.

In addition, some concurrent works have already conducted preliminary studies based on GRPO in multilingual scenarios.
\citet{park2025crosslingualcollapselanguagecentricfoundation} find that GRPO rapidly amplifies pre-training language imbalances within just a few hundred updates, resulting in the cross-lingual collapse, and language consistency reward mitigates this drift with a large drop in accuracy.
\citet{hwang2025learngloballyspeaklocally} combine SFT and multilingual GRPO with a language-consistency reward to enhance multilingual reasoning fidelity on a geography-based multilingual factual reasoning benchmark.
\citet{lee2025makingqwen3thinkkorean} only employ a customized GRPO to improve the reasoning performance on Korean.
Differently, we use the strict LC reward to achieve better input-output language consistency and design a novel CTA reward that transfers reasoning capability from English to other languages to improve the multilingual reasoning performance.




\section{Methodology}

In this section, we first briefly introduce the GRPO algorithm (\S\ref{sec:grpo}), and then present our designed rewards (\S\ref{sec:rewards}), which quantify the language consistency and alignment ratio to the English thinking sequence, besides format and answer accuracy.
Finally, we introduce our training procedure (\S\ref{sec:training-procedures}).

\subsection{Background: GRPO}\label{sec:grpo}

Recently, GRPO \cite{shao2024deepseekmathpushinglimitsmathematical} has been widely utilized for enhancing the performance of language models \cite{deepseekai2025deepseekr1incentivizingreasoningcapability, 2025arXiv250610910M, wang2025deeptransdeepreasoningtranslation, wang2025extransmultilingualdeepreasoning}.
GRPO discards the critic model and estimates the baseline from group scores instead to largely save the training costs.
Specifically, for each question $q$ in the question set $Q$, GRPO first utilizes the old policy model $\pi_{\theta_{old}}$ to samples a group of outputs $\{o_1, o_2, \cdots, o_N\}$ and then optimizes the policy model $\pi_{\theta}$ by maximizing the following objective:
\begin{equation}\label{eq:grpo}
\small
\begin{split}
\mathcal{J}_{\text{GRPO}}(\theta)\!= & \mathbb{E}{[q \sim P(Q), \{o_i\}_{i=1}^N \sim \pi_{\theta_{old}}(O|q)]} \\
&\frac{1}{N}\sum_{i=1}^{N} (\operatorname{min} 
( \frac{\pi_\theta(o_i \!\mid\! q)}{\pi_{\theta_{\text{old}}}(o_i \!\mid \!q)} A_i, \\
& \operatorname{clip}( \frac{\pi_\theta(o_i \!\mid\! q)}{\pi_{\theta_{\text{old}}}(o_i \!\mid\! q)}, 1\!-\!\epsilon, 1 \!+\! \epsilon ) A_i ) 
\!-\! \beta \mathbb{D}_{\mathrm{KL}}
),
\end{split}
\end{equation}
where $\epsilon$ and $\beta$ are hyper-parameters, and $A_i$ is the advantage computed using a group of rewards $\{r_1, r_2, \ldots, r_N\}$ corresponding to the outputs within each group:
\begin{equation}\label{eq:advantages}
\small
    A_i = \frac{r_i - {\operatorname{mean}(\{r_1, r_2, \cdots, r_N\})}}{{\operatorname{std}(\{r_1, r_2, \cdots, r_N\})}},
\end{equation}
where $r_i = R(o_i)$ is calculated by the reward function $R(o)$.

\subsection{Reward Modeling}\label{sec:rewards}
To make LRMs generate correct thinking processes and answer sequences in the input language when processing non-English inputs, we employ the following four reward modeling functions.

\paragraph{Language Consistency Reward.}
To improve the input-output language consistency, we design the LC reward to judge whether the thinking sequence $o_t$ and the answer sequence $o_a$ of the output $o$ are generated with the input language $\ell$.
First, we identify the involved language(s) of one sequence $x$ using the \texttt{langdetect}\footnote{\url{https://pypi.org/project/langdetect/}} library following \citet{wang2025polymath}. 
Formally, we define the detected language(s) set in the sequence $x$ as $\phi(x)$, and $x$ is language-consistent with $\ell$ when only one language is detected and the language is equal to $\ell$:
\begin{equation}\label{eq:lc}
\mathrm{LC}(x) = (|\phi(x)| = 1) \land (\ell \in \phi(x)),
\end{equation}
where $|\cdot|$ is the number of detected language(s) set and $\mathrm{LC}(x)$ is True or False.

Based on $\mathrm{LC}(x)$, the LC reward $R_{\text{lc}}(o)$ is defined as 0 when $o_t$ and $o_a$ are all language-consistent with $\ell$, and -1 otherwise:
\begin{equation}
  R_{\text{lc}}(o) = 
\begin{cases} 
0, & \text{if } \mathrm{LC}(o_{t}) \land \mathrm{LC}(o_{a}), \\
-1, & \text{otherwise} .
\end{cases}
\end{equation}
The LC reward strictly ensures that the model can generate the thinking and answering sequence in the input language $\ell$ by punishing the inconsistency phenomenon.

\paragraph{Cross-lingual Thinking Alignment Reward.}
Existing LRMs generally exhibit better performance on English compared to other languages \cite{huang2025surveylargelanguagemodels, yang2025languageimbalancedrivenrewarding, zhang2025cmalignconsistencybasedmultilingualalignment}, which motivates us to align the multilingual reasoning capacity to the English reasoning ability to further improve the answer correctness of multilingual responses.
Therefore, we design the CTA reward $R_{\text{cta}}(o)$, which represents the alignment ratio between the English thinking sequence $o_t^{en}$ and the current thinking sequence $o_t^{\ell}$:
\begin{equation}
 R_{\text{cta}}(o) = \mathrm{LLMJudge}(o_t^{\ell}, o_t^{en}) \in [0, 1].
\end{equation}
Specifically, we carefully design the judge instruction and request \texttt{DeepSeek-v3-0324} to evaluate the alignment ratio according to the overlap between intermediate results of $o_t^{en}$ and $o_t^{\ell}$.
Please refer to Appendix \ref{sec:appendix-judge-instruction} for the specific judge instruction.
The CTA reward utilizes the English thinking sequence as a reliable teacher to advance the cross-lingual alignment, further improving the correctness of the multilingual reasoning process.

\paragraph{Format Reward.} 
This reward is commonly used \cite{deepseekai2025deepseekr1incentivizingreasoningcapability, wang2025deeptransdeepreasoningtranslation, 2025arXiv250610910M} to ensure the format correctness of the generated outputs. 
Given a question $q_{\ell}$ in language $\ell$, the output $o$ generated by the old policy model $\pi_{\theta_{old}}$ must conform to the response pattern ``\text{<think>$o_{t}$</think>$o_{a}$}'', where ``\text{<think>}'' and ``\text{</think>}'' are two special tokens to split the thinking sequence ($o_{t}$) and the answer sequence ($o_{a}$).
Based on the strict pattern, we utilize the regular expression to verify the pattern correctness of $o$ and define the format reward as:
\begin{equation}
  R_{\text{format}}(o) = 
\begin{cases} 
0, & \text{if format is correct}, \\
-1, & \text{if format is incorrect}.
\end{cases}
\end{equation}

\paragraph{Accuracy Reward.}
For mathematical questions, the accuracy reward $R_{\text{acc}}(o)$ is widely utilized to verify the correctness of $o$:
\begin{equation}
  R_{\text{acc}}(o) = 
\begin{cases} 
1, & \text{if answer is correct}, \\
0, & \text{if answer is incorrect}.
\end{cases}
\end{equation}
Specifically, the final answer is extracted from inside the last ``\texttt{\textbackslash boxed\{\}}'' in $o$ and compared against the ground truth using a rule-based verifier \cite{sheng2024hybridflow}.

\paragraph{Overall Reward.}
Based on the above four rewards, we design the overall reward $R_{\text{all}}(o)$ as follows:
\begin{equation}
\small
R_{\text{all}}(o) \!=\!
\begin{cases} 
-1, \text{ if } R_{\text{format}}(o) \!=\! -1 \vee R_{\text{lc}}(o) \!=\! -1, \\
R_{\text{acc}}(o) \cdot (1 \!+ \!R_{\text{cta}}(o)), \text{ otherwise}.
\end{cases}
\end{equation}
Particularly, only when $R_{\text{format}}(o) \!=\! 0$ and $R_{\text{lc}}(o) \!=\! 0$, we then calculate the reward following $R_{\text{acc}}(o) \cdot (1 \!+ \!R_{\text{cta}}(o))$.

\begin{algorithm*}[t]
  \small
  \caption{Iterative Training Procedure for M-Thinker}
  \label{alg:iterative_training}
  \textbf{Input:} Cold-started model $\pi_{\theta_0}$; Multilingual questions $\mathcal{Q_{\ell}}$; Parallel English questions $\mathcal{Q}_{en}$; Reward functions $R_{\text{format}}$, $R_{\text{acc}}$, $R_{\text{lc}}$, and $R_{all}$; Hyperparameters: outer iterations $I$, sampling candidates $N$

  \begin{algorithmic}[1]
    \STATE Let $\mathbb{I}(\cdot)$ be an indicator function that returns 1 if the condition is true, and 0 otherwise
    \FOR{iteration $i = 1, \dots, I$}
        \STATE \COMMENT{\textbf{Phase A: Data Construction with Rejection Sampling}}
        \STATE Set reference model for this iteration: $\pi_{\text{ref}} \leftarrow \pi_{\theta_{i-1}}$
        \STATE Initialize RL training dataset $\mathcal{D}_{\text{RL}}^{(i)} \leftarrow \emptyset$
        
        \FOR{each question $q_{\ell} \in \mathcal{Q}_{\ell}$ with its parallel English question $q_{en} \in \mathcal{Q}_{en}$}
            \STATE Generate $N$ candidate outputs $\{o_k^{\ell}\}_{k=1}^N \sim \pi_{\text{ref}}(\cdot|q_{\ell})$
            \STATE Define $\mathcal{O}_{\text{correct}}^{\ell} = \{ o_k^{\ell} \mid \mathbb{I}(R_{\text{format}}(o_k^{\ell})=0 \land R_{\text{lc}}(o_k^{\ell})=0 \land R_{\text{acc}}(o_k^{\ell})=1) = 1 \}$
            
            \STATE Generate $N$ English candidate outputs $\{o_k^{en}\}_{k=1}^N \sim \pi_{\text{ref}}(\cdot|q_{en})$
            \STATE Define $\mathcal{O}_{\text{correct}}^{en} = \{ o_k^{en} \mid \mathbb{I}(R_{\text{format}}(o_k^{en})=0 \land R_{\text{lc}}(o_k^{en})=0 \land R_{\text{acc}}(o_k^{en})=1) = 1 \}$
            
            \IF{$0 < |\mathcal{O}_{\text{correct}}^{\ell}| < N$}
                \STATE Randomly select one correct English output as the thinking reference: ${o^{en}}^* \leftarrow \text{RandomSample}(\mathcal{O}_{\text{correct}}^{en})$
                \STATE Add the multilingual question to the training set: $\mathcal{D}_{\text{RL}}^{(i)} \leftarrow \mathcal{D}_{\text{RL}}^{(i)} \cup \{(q_{\ell}, {o^{en}}^*)\}$
            \ENDIF
        \ENDFOR

        \STATE \COMMENT{\textbf{Phase B: GRPO Training}}
        \STATE Train with GRPO (using $R_{\text{all}}$) on $\mathcal{D}_{\text{RL}}^{(i)}$ following Eq.(\ref{eq:grpo}) and update $\pi_{\theta_{i}} \leftarrow \pi_{\theta_{i-1}}$      
    \ENDFOR
  \end{algorithmic}
  \textbf{Output:} The final trained model $\pi_{\theta_I}$.
\end{algorithm*}


\subsection{Training Procedure}\label{sec:training-procedures}
We present our training procedure in Algorithm \ref{alg:iterative_training}, incorporating cold-start SFT \cite{wang2025deeptransdeepreasoningtranslation}, rejection sampling \cite{liu2024statisticalrejectionsamplingimproves}, and iterative RL training \cite{yang2025languageimbalancedrivenrewarding}. 
Specifically, given the model $\pi_{\theta}$, we first conduct the cold-start SFT to ensure that the initial model $\pi_{\theta_0}$ can generate valid samples during the GRPO training process, which is a prerequisite for effective training.
Subsequently, the model enters an iterative RL training loop. 

In each iteration $i$, we first construct the training data. 
Using the previous model $\pi_{\theta_{i-1}}$, we apply a rejection sampling strategy to select ``hard'' but solvable problems. 
Specifically, a multilingual question $q_{\ell}$ is selected if the model generates both correct and incorrect answers for it (i.e., $0 < |\mathcal{O}_{\text{correct}}^{\ell}| < N$). 
For each selected question, we also select a high-quality English output, ${o^{en}}^*$, by randomly sampling from the correct outputs for its parallel English question $q_{{en}}$. 
The thinking sequence $o^{en}_{t}$ of ${o^{en}}^*$ is used for $R_{\text{cta}}$.
The reason why we utilize the self-generated English thinking as the reference of $R_{\text{cta}}$ is that they not only do not request other models but also may have a smaller gap between the ability of non-English languages and English compared to external models.
These selected questions and their corresponding English answers form the training data $\mathcal{D}_{\text{RL}}^{(i)}$ for the current iteration.
Next, we perform GRPO training with our designed reward $R_{\text{all}}(o)$. 
The model $\pi_{\theta_{i-1}}$ is updated to $\pi_{\theta_i}$ by optimizing the GRPO objective following Eq.(\ref{eq:grpo}) on $\mathcal{D}_{\text{RL}}^{(i)}$. 
And we utilize our designed reward $R_{\text{all}}(o)$ to calculate the rewards in Eq.(\ref{eq:advantages}).
The iterative cycle of data construction and policy optimization enables the model to progressively master complex multilingual reasoning.

\section{Experiments}

\subsection{Experimental Setups}

\paragraph{Backbones and Languages.}
We select three commonly-used reasoning models with different sizes as our backbones: DeepSeek-R1-Distill-Qwen-1.5/7B \cite{deepseekai2025deepseekr1incentivizingreasoningcapability} and Qwen3-4B-Thinking-2507 \cite{yang2025qwen3technicalreport}.
The three models exhibit imbalanced reasoning performance in different languages, showing better ability in English compared to other languages.
Based on the imbalanced ability and the included languages of the MMATH \cite{luo2025mmathmultilingualbenchmarkmathematical} benchmark, we select Japanese (\textit{ja}), Korean (\textit{ko}), French (\textit{fr}), Portuguese (\textit{pt}), and Thai (\textit{th}) as the training (in-domain, ID) languages and English (\textit{en}), Spanish (\textit{es}), Arabic (\textit{ar}), Vietnamese (\textit{vi}), and Chinese (\textit{zh}) as out-of-domain (OOD) languages to observe the generalization\footnote{Since the original model performs well on \textit{en}, we actually want to observe the catastrophic forgetting phenomenon for \textit{en}. To simplify writing, we refer to it as generalization here.} of each method.
The details for each language are introduced in Table \ref{table:languages} of Appendix \ref{sec:appendix-languages}.

\paragraph{Benchmarks and Metrics.}
In this paper, we focus on the math reasoning task, which has sufficient multilingual benchmarks.
We mainly evaluate the multilingual reasoning ability on the MMATH \cite{luo2025mmathmultilingualbenchmarkmathematical} benchmark, which comprises 374 mixed-difficulty math problems sourced from AIME24/25, CNMO, and MATH-500 \cite{lightman2023letsverifystepstep}, and covers the above mentioned ten languages (\textit{ja}/\textit{ko}/\textit{fr}/\textit{pt}/\textit{th}/\textit{en}/\textit{es}/\textit{ar}/\textit{vi}/\textit{zh}).
Following \citet{luo2025mmathmultilingualbenchmarkmathematical}, we conduct each evaluation four times and report the average result across all runs. 
Specifically, for each individual evaluation, we compute the macro-average metric rather than the micro-average to account for the varying difficulty levels across subsets in MMATH.

To evaluate both the language consistency and answer accuracy of model responses, we adopt three metrics: Language Consistency (LC), Accuracy (Acc), and Language Consistency \& Accuracy (LC\&Acc).
LC assesses whether the language used throughout the response (including both the thinking and answer sequences) matches the language of the input question, referring to Eq.(\ref{eq:lc}). 
Acc measures the correctness of the final extracted answer\footnote{We directly utilize the extraction and verification tool of MMATH \cite{luo2025mmathmultilingualbenchmarkmathematical}.}, regardless of the language in which the response is generated.
LC\&Acc evaluates answer correctness only when the response \textit{o} is fully in the input language, \textit{i.e.}, $R_{\text{lc}}(o)\!=\!0 \land R_{\text{acc}}(o)=1$, which combines both language consistency and answer accuracy as our main evaluation metric.
Furthermore, we also evaluate our model on the PolyMath \cite{wang2025polymath} benchmark for additional validation. The evaluation details on PolyMath are present in Appendix \ref{sec:appendix-polymath-details}.


\paragraph{Data.}
We conduct our experiments based on the Light-R1-SFTData\footnote{\url{https://huggingface.co/datasets/qihoo360/Light-R1-SFTData}} dataset \cite{wen2025lightr1curriculumsftdpo}, which contains about 76K carefully selected data samples, \textit{i.e.}, each English question with the accurate response generated from DeepSeek-R1 \cite{deepseekai2025deepseekr1incentivizingreasoningcapability}. 
To obtain the multilingual questions, we deploy the \texttt{DeepSeek-V3-0324} model \cite{deepseekai2024deepseekv3technicalreport} to translate\footnote{The translation prompt follows \citet{wang2024ultralink} and \citet{zhang2025cmalignconsistencybasedmultilingualalignment}.} the English questions to \textit{ja}/\textit{ko}/\textit{fr}/\textit{pt}/\textit{th}.
For the cold-start SFT, we randomly sample 7.5K questions for each language and deploy the \texttt{DeepSeek-R1-0528} model \cite{deepseekai2025deepseekr1incentivizingreasoningcapability} to generate responses in the input language. 
We then filter these samples based on their LC\&Acc scores (retaining only those responses that are both language consistent with the input and answer correct) to construct the training dataset for the cold-start SFT, which comprises approximately 20K samples across all five ID languages.
For each iteration of RL training, we apply rejection sampling on the remaining data from Light-R1-SFTData. 
And we set the sampling candidates N is 8.
From the filtered RL dataset, we randomly select 3K samples per ID language for RL training.

\paragraph{Implementation Details.}
We set the iterations for RL training $I$ is 2.
The detailed training settings of cold-start SFT and iterative RL training, and generation configs are listed in Appendix \ref{sec:appendix-implementation-details}.

\begin{table*}[t]
    \centering
    \resizebox*{\linewidth}{!}{
    \begin{tabular}{l|ccccc|c|ccccc|c|c}
    \bottomrule
& \multicolumn{6}{c|}{\textbf{In-Domain Languages}} & \multicolumn{6}{c|}{\textbf{Out-of-Domain Languages}} & \\
\hline
\textbf{Methods} &	\textbf{ja}	&	\textbf{ko}	&	\textbf{fr}	&	\textbf{pt}	&	\textbf{th}	&	\textbf{\textit{ID-AVG}}	&	\textbf{en}	&	\textbf{es}	&	\textbf{ar}	&	\textbf{vi}	&	\textbf{zh}	&	\textbf{\textit{OOD-AVG}}	&	\textbf{\textit{ALL-AVG}}	\\

\toprule
\multicolumn{14}{c}{\textbf{\textit{Metric:} Language Consistency (LC, \%)}} \\
\bottomrule

\rowcolor{gray!20}
\textbf{DeepSeek-R1-Distill-Qwen-7B} & 9.49  & 	2.47  & 	16.56  & 	10.88  & 	2.19  & 	8.32  &  96.35  & 	15.61  & 	7.70  & 	23.35  & 	71.23  & 	42.85  & 	25.58  \\

\textbf{Prompt-Control (No Training)}  & 29.63  & 	2.99  & 	26.08  & 	33.77  & 	9.93  & 	20.48 &  95.47  & 	43.15  & 	8.92  & 	44.92  & 	73.58  & 	53.21  & 	36.84 \\
\textbf{DIT (No Training)}  &  68.99 & 	2.85 	& 78.28 & 	66.39 	& 15.66 & 	46.43 &  95.93 & 	66.78 & 	6.22 	& 65.79 & 	71.14 & 	61.17 & 	53.80 \\
\textbf{QRT (No Training)}  &  29.77 & 	4.21 & 	85.00 & 	67.72 & 	37.26 & 	44.79&   95.38 & 	69.07 & 	9.26 & 	62.30 & 	77.02 & 	62.61 & 	53.70 \\

\textbf{Cold-Start SFT}  & 13.69 	 & 0.64  & 	30.59  & 	21.47  & 	4.13  & 	14.10  &  \underline{98.09}  & 	28.51  & 	2.03  & 	29.81 	 & 84.87  & 	48.66 	 & 31.38  \\
\textbf{Naive-RL}  & 0.00 & 	0.00 & 	0.00 	& 0.00 & 	0.00 & 	0.00  & 96.29 & 	0.00 & 	0.00 & 	0.00 & 	85.86 & 	36.43 	& 18.22  \\
\textbf{SLC-RL}  & 91.20  & 	0.00  & 	99.54 	 & 99.09 	 & 90.18 	 & 76.00  & 99.77 	 & 99.15 	 & 1.61 	 & 81.84  & 	88.82  & 	74.24  & 	75.12  \\

\textbf{M-Thinker-7B $\Rightarrow$ Iter-1 (Ours)} & \textbf{{98.32}} 	& \underline{98.74} & 	\textbf{99.96} & 	\textbf{99.88} & 	\textbf{{99.27}} & 	\textbf{{99.23}} &  	\textbf{100.00} & 	\textbf{99.80}	& \textbf{84.68} 	& \underline{99.56} & 	\underline{89.17} 	& \textbf{94.64} & 	\textbf{96.94}  \\
\textbf{M-Thinker-7B $\Rightarrow$ Iter-2 (Ours)} & 	\underline{97.86} & 	\textbf{99.37} & 	\underline{99.50} & 	\underline{99.05} 	& \underline{95.68} 	& \underline{98.29} &  98.00 	& \underline{99.44} 	& \underline{75.12} & 	\textbf{100.00} & 	\textbf{90.97} 	& \underline{92.70} 	& \underline{95.50} \\

\toprule
\multicolumn{14}{c}{\textbf{\textit{Metric:} Accuracy (Acc, \%)}} \\
\bottomrule

\rowcolor{gray!20}
\textbf{DeepSeek-R1-Distill-Qwen-7B}  & 53.44  & 	61.61  & 	64.47  & 	62.67  & 	50.71  & 	58.58  &  65.20  & 	61.31  & 	55.28  & 	58.10  & 	52.99  & 	58.58  & 	58.58 	\\

\textbf{Prompt-Control (No Training)}  & 40.63  & 	60.18  & 	60.92  & 	58.43  & 	49.66  & 53.96  & 62.18 	 & 57.64  & 	52.24  & 	50.80  & 	57.69 	 & 56.11 	 & 55.04 	\\

\textbf{DIT (No Training)}  &  21.36 & 	40.86 	& 47.35 & 	55.72 & 	41.60 & 	41.38 &  64.51 & 	51.37 &  	50.78 & 	38.39 & 	56.98 & 	52.41 & 	46.89 \\
\textbf{QRT (No Training)}  &  30.88 & 	42.52 & 	53.92 & 	52.80 	& 32.27 & 	42.48 &  63.36 & 	54.73 	& 51.47 & 	44.79 & 	56.18 & 	54.11 & 	48.29 \\

\textbf{Cold-Start SFT}   &  48.15  & 	55.40  & 	60.78  & 	61.16  & 	49.15  & 	54.93  &  63.62  & 	61.21  & 	52.69  & 	51.76  & 	58.20  & 	57.50  & 	56.21 	\\
\textbf{Naive-RL} &  \textbf{66.11}  & 	\underline{65.18}  & 	 \underline{65.71}  & 	\underline{66.81}  & 	\textbf{65.82}  & 	\textbf{65.93}  &  \underline{69.21}  & 	\underline{64.16} 	 & \textbf{63.29}  & 	\textbf{64.42}  & 	63.60  & 	\underline{64.94}  & 	\textbf{65.43} 	\\
\textbf{SLC-RL}  & 47.00  & 	\textbf{66.86}  & 	57.91  & 	61.48  & 	49.96  & 	56.64  &  67.62  & 	61.86 	 & 60.99  & 	51.09  & 	61.17  & 	60.55  & 	58.59 \\

\textbf{M-Thinker-7B $\Rightarrow$ Iter-1 (Ours)} & 	53.92  & 	52.24  & 	60.56  & 	64.46  & 	54.71  & 	57.18  &   67.94 	 & 60.76  & 	54.79  & 	55.40  & 	\underline{63.97} 	 & 60.57  & 	58.87 	\\
\textbf{M-Thinker-7B $\Rightarrow$ Iter-2 (Ours)} &   \underline{58.23} & 	60.56 & 	\textbf{68.58} & 	\textbf{66.99} & 	 \underline{63.98} & 	 \underline{63.66} &  \textbf{71.75} & 	\textbf{68.34} & 	\underline{63.00} & 	\underline{61.72} & 	\textbf{67.25} 	& \textbf{66.41} & 	\underline{65.04}  \\

\toprule
\multicolumn{14}{c}{\textbf{\textit{Metric:} Language Consistency \& Accuracy (LC\&Acc, \%)}} \\
\bottomrule

\rowcolor{gray!20}
\textbf{DeepSeek-R1-Distill-Qwen-7B}  &  6.73  & 	2.11  & 	13.99 	 & 9.93 	 & 1.67  & 	6.89   & 65.14  & 	14.16  & 	5.47  & 	15.69  & 	45.00  & 	29.09  & 	17.99 	\\

\textbf{Prompt-Control (No Training)} & 14.62  & 	2.67  & 	20.36 	 & 26.75  & 	7.47  & 	14.37  &  61.81  & 	33.95  & 	6.79  & 	24.64  & 	46.95  & 	34.83  & 	24.60  \\

\textbf{DIT (No Training)}  & 17.99 	& 2.07 & 	43.76 & 	44.94 & 	12.55 & 	24.26 &  64.45 	& 45.90 & 	4.15 & 	35.13 	& 48.24 	& 39.57 & 	31.92  \\
\textbf{QRT (No Training)}   &  18.51 	& 3.82 & 	52.17 & 	44.66 & 	18.02 & 	27.44 &  63.26 & 	48.69 & 	6.98 & 	39.82 & 	51.30 & 	42.01 & 	34.72 \\

\textbf{Cold-Start SFT}  & 8.58  & 	0.44  & 	23.64  & 	18.51  & 	2.13  & 	10.66  &  63.58  & 	25.22  & 	1.41  & 	20.03  & 	50.50  & 	32.15 	 & 21.40 	\\
\textbf{Naive-RL}  & 0.00  & 	0.00  & 	0.00  & 	0.00  & 	0.00  & 	0.00   & \underline{68.48}  & 	0.00  & 	0.00  & 	0.00  & 	54.11  & 	24.52  & 	12.26 \\
\textbf{SLC-RL}  & 46.52  & 	0.00  & 	57.87  & 	61.42  & 	49.90  & 	43.14  &  67.60 	 & \underline{61.70}  & 	1.57 	 & 49.57  & 	53.96  & 	46.88  & 	45.01 \\

\textbf{M-Thinker-7B $\Rightarrow$ Iter-1 (Ours)}	 & \underline{53.30} 	 & \underline{52.12}  & 	\underline{60.54} 	 & \underline{64.34}  & \underline{54.71} 	 & \underline{57.00}  &  67.94  & 	60.58 	 & \underline{52.14}  & 	\underline{55.38}  & 	\underline{56.21} 	 & \underline{58.45}  & 	\underline{57.73}  \\
\textbf{M-Thinker-7B $\Rightarrow$ Iter-2 (Ours)}	 & \textbf{57.50}  & 	\textbf{60.26}  & 	\textbf{68.52}  & 	\textbf{66.87}  & 	\textbf{63.44}  & 	\textbf{63.32}  &  \textbf{71.71} 	 & \textbf{68.22}  & 	\textbf{53.70}  & 	\textbf{61.72}  & 	\textbf{60.58}  & 	\textbf{63.18} 	 & \textbf{63.25}  \\
\toprule

    \end{tabular}
    }
    \caption{
        The LC, Acc, and LC\&Acc (\%) results on the MMATH benchmark of the DeepSeek-R1-Distill-Qwen-7B backbone. ``\textit{\textbf{ID-avg}}/\textit{\textbf{OOD-avg}}'' is the average result of five In-Domain/Out-of-Domain languages and ``\textit{\textbf{ALL-AVG}}'' is the average result of all ten languages. The result in \textbf{bold} means the best result, and the \underline{underlined} result means the second-best result in each setting.
        ``\textbf{Iter-1/2}'' means the training iteration 1/2.
    }
    \label{table:main-res-7B}
\end{table*}

\subsection{Baselines}
\paragraph{Prompt-Control.}
Following \citet{wang2025polymath}, we concatenate the language control instructions after the input prompts to make the model generate responses using the same language as the query.
Please refer to Figure \ref{fig:prompt-control-instructions} of Appendix \ref{sec:appendix-baselines-instructions} for the detailed language control instructions of each language.

\paragraph{DIT.} 
Discourse-Initiated Thinking \cite{luo2025mmathmultilingualbenchmarkmathematical} appends the most popular beginning discourse markers in each language after the ``<think>'' token, encouraging models to initiate their reasoning using multilingual discourse cues as entry points into the thinking process. 
The used multilingual discourse marks are shown in Figure \ref{fig:dit-instructions} of Appendix \ref{sec:appendix-baselines-instructions}.

\paragraph{QRT.}
Question-Restatement Thinking \cite{luo2025mmathmultilingualbenchmarkmathematical} restates the question in the target language at the beginning of the thinking process, which encourages the model to think in the target language.
The restatement instructions for each language are listed in Figure \ref{fig:qrt-instructions} of Appendix \ref{sec:appendix-baselines-instructions}.

\paragraph{Cold-Start SFT.}
We conduct the cold-start SFT training on the constructed training dataset.

\paragraph{Naive-RL.}
We equip the GRPO algorithm only with the accuracy reward to conduct the RL training based on the same cold-started SFT model. The training dataset is the same as our first training iteration (Iter-1).

\paragraph{SLC-RL.}
We equip the GRPO algorithm with the accuracy reward and a soft language consistency reward \cite{2025arXiv250610910M} to conduct the RL training, {i.e.}, $R(o) = R_{\text{format}}(o)*(R_{\text{acc}}(o) + R_{\text{slc}}(o))$.
When the format is correct: $R_{\text{format}}(o)=1$, when the answer is correct: $R_{\text{acc}}(o)=0.9$, and when the language is consistent with the input language: $R_{\text{slc}}(o) = 0.1$, otherwise, $R_{\text{format}}(o)=R_{\text{acc}}(o)=R_{\text{slc}}(o)=0$.
The initial policy model (after cold-start SFT) and training dataset is the same as our first training iteration (Iter-1).



\begin{table*}[h]
    \centering
    \resizebox*{\linewidth}{!}{
    \begin{tabular}{l|ccc|ccc|ccc}
    \bottomrule
     & \multicolumn{3}{c|}{\textbf{LC}} & \multicolumn{3}{c|}{\textbf{Acc}} & \multicolumn{3}{c}{\textbf{LC\&Acc}}\\
    \hline
    \textbf{Methods} & \textbf{\textit{ID-AVG}} & \textbf{\textit{OOD-AVG}} & \textbf{\textit{ALL-AVG}} & \textbf{\textit{ID-AVG}} & \textbf{\textit{OOD-AVG}} & \textbf{\textit{ALL-AVG}} & \textbf{\textit{ID-AVG}} & \textbf{\textit{OOD-AVG}} & \textbf{\textit{ALL-AVG}} \\
    \hline
\rowcolor{gray!20}
\textbf{DeepSeek-R1-Distill-Qwen-1.5B} &  5.98 &	36.11 &	21.04 & 34.81 &	39.83 &	37.32 & 3.87 &	19.19 &	11.53 \\
Prompt-Control (No Training) & 12.64 	& 48.52 &	30.58 &  31.95 &	33.55 &	32.75  & 5.65 &	22.32 	& 13.99 \\
DIT (No Training) & 22.48 	& 43.30 	& 32.89  & 23.10 	& 28.50 	& 25.80  & 11.56 	& 23.68 	& 17.62  \\
QRT (No Training) & 23.47 	& 45.46 	& 34.46  & 19.26 	& 28.11 	& 23.69  & 11.68 	& 23.87 	& 17.78  \\
Cold-Start SFT  & 23.73 	& 47.75 	& 35.74  & 19.18 	& 27.79 	& 23.49  & 7.39 	& 21.73 	& 14.56 \\
Naive-RL & 0.00 	& 30.97 	& 15.48 &  \textbf{49.99} & 	\textbf{50.08} 	& \textbf{50.04} &  0.00 & 	16.16 	& 8.08 \\
SLC-RL  & 0.00 	& 37.16 	& 18.58 &  \underline{46.80} 	& \underline{49.16} 	& \underline{47.98} &  0.00 & 	19.47 	& 9.74  \\
\textbf{M-Thinker-1.5B $\Rightarrow$ Iter-1 (Ours)} &  \underline{99.19} & \textbf{84.48} & \textbf{91.83} &  35.59 & 44.22 & 39.90 & \underline{35.39} & \underline{38.37} & \underline{36.88} \\
\textbf{M-Thinker-1.5B $\Rightarrow$ Iter-2 (Ours)} &  \textbf{99.49} & \underline{79.51} & \underline{89.50} &  42.72 & 46.53 & 44.62 & \textbf{42.47} & \textbf{38.83} & \textbf{40.65} \\

    \toprule
    \end{tabular}
    }
    \caption{
    The LC, Acc, and LC\&Acc (\%) results on the MMATH benchmark of the DeepSeek-R1-Distill-Qwen-1.5B backbone. The detailed results for each language are list in Table \ref{table:main-res-1.5B} of Appendix \ref{sec:appendix-res-qwen1.5}.
    }
    \label{table:main-results-1.5b2}
\end{table*}

\begin{table*}[h]
    \centering
    \resizebox*{\linewidth}{!}{
    \begin{tabular}{l|ccc|ccc|ccc}
    \bottomrule
     & \multicolumn{3}{c|}{\textbf{LC}} & \multicolumn{3}{c|}{\textbf{Acc}} & \multicolumn{3}{c}{\textbf{LC\&Acc}}\\
    \hline
    \textbf{Settings} & \textbf{\textit{ID-AVG}} & \textbf{\textit{OOD-AVG}} & \textbf{\textit{ALL-AVG}} & \textbf{\textit{ID-AVG}} & \textbf{\textit{OOD-AVG}} & \textbf{\textit{ALL-AVG}} & \textbf{\textit{ID-AVG}} & \textbf{\textit{OOD-AVG}} & \textbf{\textit{ALL-AVG}} \\
    \hline
\rowcolor{gray!35}
\textbf{M-Thinker-1.5B $\Rightarrow$ Iter-1 (Ours)} &  99.19 & 84.48 & 91.83 &  35.59 & 44.22 & 39.90 & 35.39 & 38.37 & 36.88 \\
\textbf{\quad \quad  \textit{w/o} $R_{\text{cta}}$} & 99.16 & 92.44 & 95.80 & 31.72 & 39.85 & 35.78 & 31.68 & 37.18 & 34.43 \\
\textbf{\quad \quad \textit{w/o} $R_{\text{lc}}$} & 0.00 & 35.61 & 17.80 & 50.22 & 50.83 & 50.52 & 0.00 & 18.66 & 9.33 \\
\textbf{\quad \quad \textit{w/o} ($R_{\text{cta}}$ \& $R_{\text{lc}}$)} & 0.00  & 	30.97  & 	15.48  & 49.99  & 	50.08 	 & 50.04 & 0.00 	 & 16.16 	 & 8.08  \\

\textbf{\quad \quad \textit{w/o} Cold-Start SFT}  & 99.19  & 	84.33  & 	91.76  & 33.60 	 & 42.83  & 	38.22  & 33.35  & 	36.91  & 	35.13  \\
\textbf{\quad \quad \textit{w/o} Rejection Sampling} & 99.71  & 	85.31 	 & 92.51  & 33.87 	 & 41.24  & 	37.55  & 33.73  & 	35.48 	 & 34.60  \\
\textbf{\quad \quad \textit{w/} $o_t^{en}$ from Light-R1 for $R_{\text{cta}}$} & 99.76 & 88.27 & 94.01 & 33.71 & 41.87 & 37.79 & 33.67 & 37.65 & 35.66 \\
    \toprule
    \end{tabular}
    }
    \caption{
        The ablation results of the MMATH benchmark based on our M-Thinker-1.5B (Iter-1). ``\textit{w/o}'' means without one setting and ``\textit{w/}'' means with one setting.
    }
    \label{table:ablation-results}
\end{table*}

\subsection{Main Results}
\paragraph{Performance of our M-Thinker.} We report the evaluation results on MMATH of the three backbones in Table \ref{table:main-res-7B}, Table \ref{table:main-results-1.5b2}, and Table \ref{table:main-res-4B} (Appendix \ref{sec:appendix-qwen3}).
The results demonstrate that our M-Thinker-1.5B/4B/7B achieves excellent improvement on LC, Acc, and the combined metric (LC\&Acc). 
On the main evaluation metric (LC\&Acc), our M-Thinker-1.5B/4B/7B (Iter-1) drastically outperforms all baselines, which highlights the effectiveness of our designed rewards in simultaneously optimizing for correctness and language fidelity.
Furthermore, our M-Thinker-1.5B/7B (Iter-2) achieves further improvement than Iter-1, which proves that our iterative training procedure can progressively enhance the model's capabilities.
And the performance on LC\&Acc of our M-Thinker-1.5B/7B (Iter-2) has surpassed the performance on Acc of the backbones DeepSeek-R1-Distill-Qwen-1.5B/7B, which means that responding in the input language can exceed the performance of responding in English or other default languages.
This superior performance indicates that our method mitigates the trade-off between language consistency and answer accuracy, achieving powerful multilingual reasoning ability.

\paragraph{Performance of baselines.} No training baselines have a minor improvement on LC\&Acc, and QRT outperforms DIT and Prompt-Control.
The performance of these prompt-based methods heavily depends on the original instruction-following ability of backbones, \textit{i.e.}, the larger improvement on 7B than 1.5B.
Additionally, the improvement on LC and the decrease on Acc also reflect the trade-off between the language consistency and answer accuracy.
Naive-RL (GRPO only with the accuracy reward) shows the best results on Acc but the lowest LC (0.0) since the responses generated in English can obtain a higher reward score during RL training, so that the trained model is most likely to think and answer in English, which is contrary to the goal of a multilingual reasoning model.
Although SLC-RL is trained with a soft language consistency reward, the models still struggle to maintain language consistency, particularly for the 1.5B backbone.
By contrast, our method with the strict LC reward can promote the input-output language consistency while having no degradation\footnote{More detailed analyses about hard/soft LC reward are listed in Appendix \ref{sec:appendix-hard-lc}.} on Acc compared to SLC-RL\footnote{We also conduct SLC-RL with the same reward magnitude as ours and present it in Table \ref{table:slc-results} of Appendix \ref{sec:appendix-soft-lc}.} (58.87 vs. 58.59).

\paragraph{OOD generalization.} Refer to the ``\textit{{OOD-avg}}'', our M-Thinker also significantly surpasses other baselines, which indicates that the reasoning patterns learned through our rewards and training procedure are not confined to the training languages but are successfully transferred to unseen languages.
The evaluation results on PolyMath (as shown in Table \ref{table:main-res-polymath} in Appendix \ref{sec:appendix-polymath}) also present similar trends, which further prove the superiority of our method.

In summary, these results demonstrate that our M-Thinker effectively improves both the language consistency and answer accuracy in multilingual reasoning scenarios.

\section{Analysis}

\subsection{Ablation Study}
We conduct an ablation study to verify the effectiveness of our designed reward functions and involved training strategies.
The ablation results listed in Table \ref{table:ablation-results} show that the LC\&Acc performance degrades in both ID and OOD languages without $R_{\text{cta}}$.
For the setting ``\textit{w/o} $R_{\text{lc}}$'', although the Acc improves over M-Thinker-1.5B, the model responds to all questions in English, resulting in the lowest language consistency.
``\textit{w/o} ($R_{\text{cta}}$ \& $R_{\text{lc}}$)'' present the lowest performance.
These results prove the effectiveness of our designed reward functions.
Additionally, ``\textit{w/o} Cold-Start SFT'' and ``\textit{w/o} Rejection Sampling'' also have a performance decline, which demonstrates the necessity of these strategies.
Furthermore, directly using English responses from the Light-R1-SFT dataset (which is generated by DeepSeek-R1) for $R_{\text{cta}}$ also underperforms our M-Thinker (using generated English responses from the model itself), since the latter may have a smaller gap between the abilities of non-English languages and English.
Detailed results of each ablation setting are listed in Table \ref{table:ablation-detailed-res} of Appendix \ref{sec:appendix-detailed-ablation-res}.

\subsection{Effects of Different Judge Models for $R_{\text{cta}}$}
In this section, we analyze the effects of different judge models for calculating $R_{\text{cta}}$ on performance and report the results in Table \ref{table:judge-model-analysis}. \\
\textbf{Findings 1: Frontier small LLMs can also provide reliable rewards.} Beyond DeepSeek-V3, we also test two smaller models as the judge model, i.e., Qwen3-30B-A3B and Qwen3-4B.
Although smaller, these two frontier models still deliver notable performance gains while being more cost-efficient than DeepSeek.
Besides, we also try another small model, Qwen2.5-7B-Instruct, that is relatively outdated compared to other models. 
We find that it decreases the overall performance (32.91\%) compared to ``\textit{w/o} $R_{\rm cta}$'' due to the limited multilingual capability, demonstrating that the multilingual capability of the judge model is crucial for the effectiveness of the $R_{\text{cta}}$ reward. \\
\textbf{Findings 2: $R_{\text{cta}}$ achieves cross-lingual transfer of the reasoning capability from English to other languages.}
As shown in Table \ref{table:judge-model-analysis}, we also find that $R_{\text{cta}}$ significantly brings the accuracy gap between English and other languages with multiple judge models according to the ``\textit{GAP}'' values.
This further proves the effectiveness of $R_{\text{cta}}$ for bridging the multilingual reasoning gap in existing models.

\begin{table}[t]
    \centering
    \resizebox*{\linewidth}{!}{
    \begin{tabular}{l|ccc|c}
    \bottomrule

\textbf{Judge Models} &	\textbf{\textit{ID-AVG}}	&	\textbf{\textit{OOD-AVG}} & \textbf{\textit{ALL-AVG}} & \textit{\textbf{GAP}}$\downarrow$\\
\hline
\textit{w/o} $R_{\text{cta}}$ & 31.68 & 37.18 & 34.43 & 13.47 \\
\hline
DeepSeek-V3-0324 & \textbf{35.39} & \textbf{38.37} & \textbf{36.88} & \textbf{9.24} \\
Qwen3-4B-Instruct-2507 &  32.48 	& 38.51 	& 35.49 & 12.56 \\
Qwen3-30B-A3B-Instruct-2507 & 33.12 &		37.08 &		35.10 & 11.75 \\
Qwen2.5-7B-Instruct & 31.69 &	34.13 &	32.91 &  13.65 \\

\toprule

    \end{tabular}
    }
    \caption{
        The LC\&Acc results of different judge models for $R_{\text{cta}}$ based on our M-Thinker-1.5B (Iter-1). ``\textit{GAP}'' denotes the accuracy gap between English and other languages.
    }
    \label{table:judge-model-analysis}
\end{table}

\begin{table}[t]
    \centering
    \resizebox*{\linewidth}{!}{
    \begin{tabular}{c|cccccccccc}
    \bottomrule

\textbf{Data} &	\textbf{ja}	&	\textbf{ko}	&	\textbf{fr}	&	\textbf{pt}	&	\textbf{th}	&	\textbf{en}	&	\textbf{es}	&	\textbf{ar}	&	\textbf{vi}	&	\textbf{zh}	 \\
\hline
1.5B & 0.22 & 	\textbf{0.02} & 	7.05 & 	11.92 & 	0.12 & 	46.56 & 	13.38 & 	0.16 & 	3.56 	& 32.30  \\
\hline
\textbf{fr} &  2.73 	& 0.00 & 	\color{blue}\textbf{37.12} &  	\textbf{34.72} & 	7.20 & 	\textbf{52.27} & 	\textbf{37.21} & 	3.54 	& 20.92 	& 38.45  \\

\textbf{ja}  & \color{blue}\textbf{26.76} & 	0.00 & 	21.91 & 	32.23 & 	\textbf{8.97} & 	49.55 	& 35.74 & 	\textbf{3.79} 	& \textbf{23.17} & 	\textbf{39.69} \\

\toprule

    \end{tabular}
    }
    \caption{
        The LC\&Acc generalization results on OOD languages when only using \textit{fr}/\textit{ja} as training data for DeepSeek-R1-Distill-Qwen-1.5B. The {\color{blue} blue} results mean the performance on the training language. The results in \textbf{bold} represent the best result in each language.
    }
    \label{table:ood-analysis}
\end{table}

\subsection{Generalization Study}
In this section, we investigate the generalization to non-training (OOD) languages when training on different languages.
Specifically, we separately use \textit{fr} and \textit{ja} to train the model and observe the performance of the other nine languages (as shown in Table \ref{table:ood-analysis}).
We find that if training on \textit{fr}, the performance of \textit{pt}, \textit{es}, and \textit{en} is better than training on \textit{ja} since \textit{pt}/\textit{es}/\textit{en} and \textit{fr} all belong to the Indo-European language family (as introduced in Table \ref{table:languages} of Appendix \ref{sec:appendix-languages}).
By contrast, training on \textit{ja} shows better generalization to \textit{zh}/\textit{vi}.
We guess that although \textit{ja} generally is regarded as an Isolate language, some scripts are sourced from Chinese, and a few scripts of Vietnamese also source from Chinese.
Additionally, since \textit{ko} is an isolate language with a writing system distinct from those of \textit{ja} and \textit{fr}, it achieves the lowest generalization (0.0).
Overall, these results indicate that if you want to improve the performance of one language, the similar or same-language-family languages must be added to the training dataset.

\subsection{Human Evaluation}

To empirically validate the reliability of our utilized judge model (DeepSeek-V3-0324) for $R_{\text{cta}}$, we conduct a human evaluation on a randomly sampled subset of the training data (30 samples per language). 
We collaborate with a professional data annotation service to recruit linguistic experts proficient in the target languages (\textit{ja}/\textit{ko}/\textit{fr}/\textit{pt}/\textit{th}). 
We request that they return the alignment ratios, and the detailed annotation guidelines are introduced in Appendix \ref{sec:appendix-human}.
We then calculate the Pearson correlation between these human-annotated ratios and the judge model's CTA scores. 
The results in Table \ref{table:human-evaluation} demonstrate a positive Pearson correlation coefficient across different languages, confirming that the judge model aligns well with human judgment and proving the reliability of the CTA reward.

\begin{table}
    \centering
    \resizebox*{\linewidth}{!}{
    \begin{tabular}{c|ccc}
    \bottomrule
        \textbf{Language} & \textbf{from Human} & \textbf{from LLM} & \textbf{Pearson Correlation}\\
    \hline
        \textbf{ja} & 0.72 & 0.69 & 0.87 \\
        \textbf{ko} & 0.68 & 0.64 & 0.86 \\
        \textbf{fr} & 0.85 & 0.81 & 0.93 \\
        \textbf{pt} & 0.82 & 0.78 & 0.91 \\
        \textbf{th} & 0.63 & 0.57 & 0.83 \\
    \hline
        \textbf{AVG} & 0.74 & 0.70 & 0.88 \\
    \toprule
    \end{tabular}}
    \caption{The Pearson correlation between human-annotated ratios and the judge model's CTA scores (DeepSeek-V3-0324).}
    \label{table:human-evaluation}
\end{table}

\section{Conclusion}
In this paper, we design a Language Consistency reward to strictly enforce input-output language consistency and a Cross-lingual Thinking Alignment reward to further improve the accuracy of multilingual answers.
Additionally, we train M-Thinker-1.5B/4B/7B models with a systematic training procedure incorporating cold-start SFT, rejection sampling, and iterative RL training.
Experimental results on the MMATH and PolyMath show that our M-Thinker models exhibit excellent multilingual reasoning performance.
In summary, our work offers an effective method and valuable empirical insights for the community to enhance the intrinsic multilingual capabilities of LRMs.

\section*{Limitations}
In this paper, we only conduct experiments on five languages (3K samples for each language) and set the RL training iterations to 2 due to time and resource limitations.
We believe that more languages, more training samples, and more RL training iterations will achieve better performance.
And we only train models of the 1.5B/4B/7B sizes due to the limited GPU resources, but we think that our designed reward functions and utilized training procedure can be applied to train models of bigger sizes.
Additionally, we utilize the \texttt{langdetect} library to detect involved languages in one sequence for the LC Reward following \citet{wang2025polymath}.
However, there are some other language detection tools or models that we do not test, such as xlm-roberta-base-language-detection \cite{conneau2020unsupervisedcrosslingualrepresentationlearning}, \texttt{Cld3}\footnote{\url{https://pypi.org/project/pycld3/}}, and \texttt{FastText}\footnote{\url{https://pypi.org/project/fasttext-langdetect/}}. 
We will try and investigate a more robust and faster language detection method in the future.
Although we have tested different-size judge models for $R_{\text{cta}}$ as a valuable selective guideline, we acknowledge that $R_{\text{cta}}$ introduces additional training overhead.
In the future, we will explore more judge models to select the most effective and efficient model for $R_{\text{cta}}$.
As for extremely low-resource languages where the Judge model completely fails to comprehend the input, the CTA reward would indeed be unreliable.
In such cases, we suggest falling back to only Format+Acc+LC reward or replacing it with rule-based CTA rewards. 
Considering our paper mainly focuses on language consistency and the effectiveness of the CTA reward on multilingual reasoning, we leave the study of extremely low-resource languages for further study.


\bibliography{custom}

\newpage
\appendix
\section{Instruction for Cross-lingual Thinking Alignment Reward}\label{sec:appendix-judge-instruction}
The designed judge instruction for requesting \texttt{DeepSeek-V3-0324} to evaluate the alignment ratio is as follows:
\begin{tcolorbox} 
\fontsize{10pt}{11pt}\selectfont
\# Task

Analyze and quantify the consistency of \textbf{key intermediate results} between an English and a [target] thought process for a given math problem.\\

\# Inputs

I will provide you with three items:
[English Math Problem]: The original problem in English.

[English Thought Process]: The step-by-step reasoning for solving the problem in English.

[[target] Thought Process]: The step-by-step reasoning for solving the problem in [target].\\

\# Instructions

You must perform the following analysis internally:

Identify \textbf{all key intermediate results} from the [English Thought Process]. Key results include variable definitions, equations, critical calculation values, and the final answer.

For each key result identified in English, find its \textbf{mathematical equivalent} in the [[target] Thought Process].

Calculate the \textbf{consistency score} using the following formula:
\textbf{Score = (Number of matched, mathematically equivalent pairs) / (Total number of key results identified in the English process)}\\

\# Output Format

Your final output MUST BE \textbf{a single decimal number between 0 and 1}. And the number should be wrapped by <score> and </score>.
Do NOT include any text, explanation, titles, analysis, or any other characters. The response must only be the number itself wrapped by <score> and </score>.\\

Example of a valid response:

<score>0.925</score>

-------

[English Math Problem]: [en-question]

[English Thought Process]: [en-think]

[[target] Thought Process]: [x-think]
\end{tcolorbox}

\section{Experimental Details}

\subsection{Introduction of Different Languages}\label{sec:appendix-languages}
The language families and writing systems \cite{zhang-etal-2025-less} of all ID/OOD languages are listed in Table \ref{table:languages}.
Specifically, \textit{fr}, \textit{pt}, and \textit{es} all belong to the Romance branch of the Indo-European family, \textit{ja} is often considered as the Isolate language, through its writing system incorporates Kanji,  which originated from \textit{zh}.

\begin{table*}[ht]
    \centering
    \resizebox*{0.9\linewidth}{!}{
    \begin{tabular}{l|ll}
    \bottomrule
        \textbf{Languages}	& \textbf{Language Family} & \textbf{Writing System} \\
        
    \hline
    English (\textit{en}) & Indo-European (Germanic) & Latin alphabet (26 letters) \\
    French (\textit{fr}) &	Indo-European (Romance)	& Latin alphabet (26 letters) \\
    Portuguese (\textit{pt}) &	Indo-European (Romance) &	Latin alphabet (26 letters + diacritics) \\
    Spanish (\textit{es}) & Indo-European (Romance) & Latin alphabet (27 letters, including ñ) \\
    Japanese (\textit{ja}) & Japonic (Isolate Language) & Japanese script (Kanji + Hiragana + Katakana) \\
    Korean (\textit{ko}) & Koreanic (Isolate Language) & Hangul (24 basic letters, often syllabically grouped) \\
    Thai (\textit{th}) & Kra–Dai (Tai) & Thai script (44 consonants + vowel symbols, abugida) \\
    Arabic (\textit{ar}) & Afro-Asiatic (Semitic) & Arabic script (28 letters, right-to-left) \\
    Vietnamese (\textit{vi}) & Austroasiatic (Vietic) & Latin alphabet (Vietnamese variant) with diacritics (29 letters) \\
    Chinese (\textit{zh}) & Sino-Tibetan (Sinitic) & Chinese characters \\

    \toprule
    \end{tabular}
    }
    \caption{
        The detailed language families and writing systems for all ID/OOD languages.
    }
    \label{table:languages}
\end{table*}

\subsection{Evaluation Details for PolyMath}
\label{sec:appendix-polymath-details}
PolyMath \cite{wang2025polymath} is a multilingual mathematical reasoning benchmark covering 18 languages and 4 easy-to-hard difficulty levels. 
In our experiments, we only test 10 languages overlapped with MMATH.
For PolyMath, we also conduct each evaluation four times and report the average result across all runs.
Differently, we report the Difficulty-Weighted Accuracy (DW-ACC) \cite{wang2025polymath}, which assigns level-specific weights $w_1, w_2, w_3, w_4$ to each problem from the low/medium/high/top level, respectively.
Specifically, the weights double at each ascending level: $w_1 \!=\! 1$, $w_2 \!=\! 2$, $w_3 \!=\! 4$, and $w_4 \!=\! 8$, which provides a more reliable measure of performance by minimizing the impact of success on easier problems and placing greater emphasis on correct answers at higher difficulty levels.
Given the accuracy at each level $a_1, a_2, a_3, a_4$, DW-ACC is defined as:
\begin{equation}
    \begin{aligned}
        \text{DW-ACC} = \frac{\sum_{i=1}^4 w_i a_i}{\sum_{i=1}^4 w_i} = \sum_{i=1}^4 \left( \frac{2^{i-1}}{15} a_i \right).
    \end{aligned}
\end{equation}
Based on DW-ACC, we also calculate and report the LC\&DW-ACC.

\subsection{Implementation Details}
\label{sec:appendix-implementation-details}

\paragraph{Cold-Start SFT.}
We use the Llama-Factory\footnote{\url{https://github.com/hiyouga/LLaMA-Factory}} framework~\cite{zheng2024llamafactory} for the cold-start SFT \cite{zhang2025dualspaceframeworkgeneralknowledge, zhang2025aligndistiltokenlevellanguagemodel, zhang2024multilingualknowledgeeditinglanguageagnostic}.
For DeepSeek-R1-Distill-Qwen-1.5B, we set the learning rate to 1e-6, the batch size to 256, and the training epoch to 1.
For DeepSeek-R1-Distill-Qwen-7B, we set the learning rate to 5e-7, the batch size to 256, and the training epoch to 1.
All SFT experiments are conducted on 1$\times$NVIDIA H20 GPUs (96G).
DeepSpeed ZeRO-2/ZeRO-3 optimization~\cite{10.1145/3394486.3406703} during SFT is adopted for DeepSeek-R1-Distill-Qwen-1.5B/7B, respectively.
Additionally, we deploy \texttt{DeepSeek-V3-0324} and \texttt{DeepSeek-R1-0528} on 2$\times$NVIDIA H20 GPU (96G) during the construction of the training dataset for the cold-start SFT.

\paragraph{RL Training.}
Following previous work \cite{deepseekai2025deepseekr1incentivizingreasoningcapability, wang2025deeptransdeepreasoningtranslation, wang2025extransmultilingualdeepreasoning}, We use GRPO algorithm implemented by verl\footnote{\url{https://github.com/volcengine/verl}}~\cite{sheng2024hybridflow}.
We conduct all RL training experiments on 8$\times$8 H20 GPUs, and we use another 2$\times$8 H20 GPUs to deploy \texttt{DeepSeek-V3-0324} to calculate the CTA reward.
For DeepSeek-R1-Distill-Qwen-1.5B/7B, we set the batch size to 512, the learning rate to 5e-6/3e-6, the rollout number to 8 and the rollout temperature to 0.9, and the KL loss coefficient to 0.0.
The number of training epochs is set to 15.
For Iter-1 and Iter-2, we set the max sequence length to 16384 and 24000, respectively.

\paragraph{Generation Details.}
During evaluation, we use the vLLM toolkit\footnote{\url{https://github.com/vllm-project/vllm}} to accelerate the model generation process.
For the original backbone and no-training baselines, we use the recommended sampling decoding strategy \cite{deepseekai2025deepseekr1incentivizingreasoningcapability} with 0.6 temperature and 0.95 top-p value.
For other training baselines, we set the sampling decoding strategy with 0.9 temperature and 0.95 top-p value for the best performance.
During the RL training, we test the checkpoints from step-320 to step-435 (per 5 steps) for the best performance.

\subsection{Instructions of No-Training Baselines}\label{sec:appendix-baselines-instructions}
\begin{figure}[h]
    \centering
    \includegraphics[width=\linewidth]{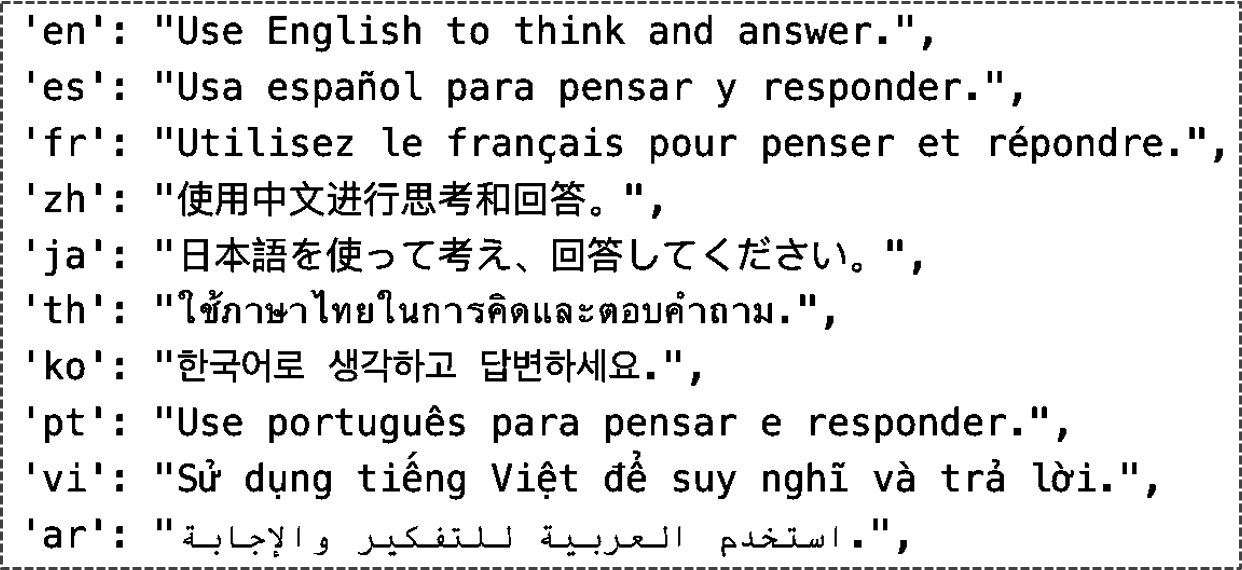}
    \caption{The language control instructions \cite{wang2025polymath} of the Prompt-Control baseline.}
    \label{fig:prompt-control-instructions}
\end{figure}

\begin{figure}[h]
    \centering
    \includegraphics[width=0.5\linewidth]{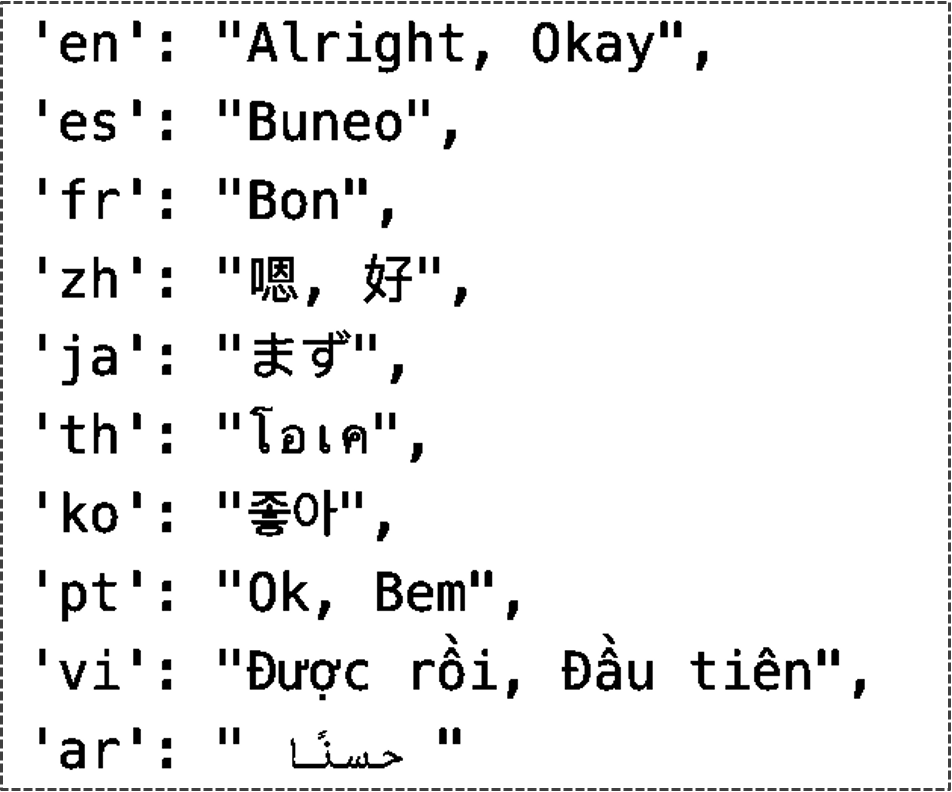}
    \caption{The multilingual discourse marks for each language \cite{luo2025mmathmultilingualbenchmarkmathematical} of the DIT baseline.}
    \label{fig:dit-instructions}
\end{figure}

\begin{figure*}[h]
    \centering
    \includegraphics[width=0.9\linewidth]{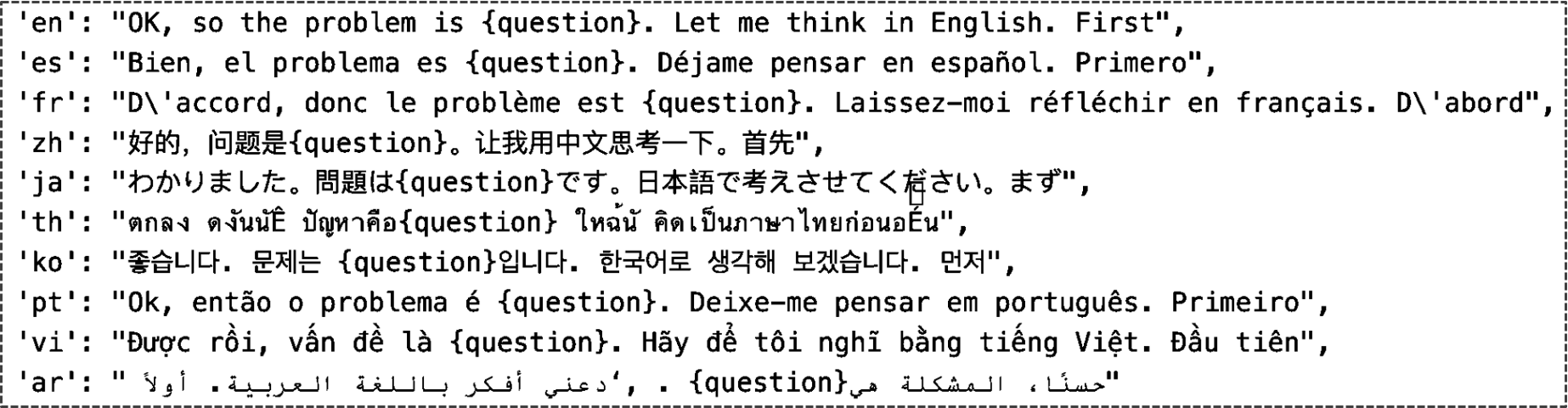}
    \caption{The restatement instructions \cite{luo2025mmathmultilingualbenchmarkmathematical} of the QRT baseline.}
    \label{fig:qrt-instructions}
\end{figure*}

We show the detailed instructions for Prompt-Control, DIT, and QRT in Figure \ref{fig:prompt-control-instructions}, \ref{fig:dit-instructions}, and \ref{fig:qrt-instructions}, respectively.

\section{Additional Results}\label{sec:appendix-results}

\begin{table*}[t]
    \centering
    \resizebox*{\linewidth}{!}{
    \begin{tabular}{l|ccccc|c|ccccc|c|c}
    \bottomrule
& \multicolumn{6}{c|}{\textbf{In-Domain Languages}} & \multicolumn{6}{c|}{\textbf{Out-of-Domain Languages}} & \\
\hline
\textbf{Methods} &	\textbf{ja}	&	\textbf{ko}	&	\textbf{fr}	&	\textbf{pt}	&	\textbf{th}	&	\textbf{\textit{ID-AVG}}	&	\textbf{en}	&	\textbf{es}	&	\textbf{ar}	&	\textbf{vi}	&	\textbf{zh}	&	\textbf{\textit{OOD-AVG}}	&	\textbf{\textit{ALL-AVG}}	\\

\toprule
\multicolumn{14}{c}{\textbf{\textit{Metric:} Language Consistency (LC, \%)}} \\
\bottomrule
\rowcolor{gray!20}

\textbf{Qwen3-4B-Thinking-2507}  &	0.00 & 	0.00 & 	0.00 	& 0.00 & 	0.00 & 	0.00  & 99.79 & 	0.00 & 	0.00 & 	0.00 & 	65.52 & 	33.06 & 	16.53 	\\

\textbf{Prompt-Control (No Training)} &	 0.00 & 	0.00 & 	0.00 	& 0.00 & 	0.00 	& 0.00  & 99.94 	& 0.00 	& 0.02 & 	0.02 & 	68.50 	& 33.70 & 	16.85 	\\
\textbf{DIT (No Training)} &  95.13 & 	16.66 & 	2.37 & 	94.31 	& 9.44 	& 43.58  & 99.94 	& 97.71 	& 64.13 	& 98.72 	& 82.61 	& 88.62 & 	66.10 \\
\textbf{QRT (No Training)} &  99.13 	& 97.17 	& 96.87 & 	92.35 & 	96.31 & 	96.37  & \textbf{99.96} 	& \textbf{99.32} & 	96.83 & 	\textbf{99.35} &  	\textbf{84.21} & 	\textbf{95.94} 	& 96.15 \\

\textbf{Cold-Start SFT} &	 68.73 	& 54.72 	& 16.64 	& 3.97 & 	17.64 & 	32.34 & 89.50 	& 9.89 	& 1.66 	& 24.18 & 	64.67 & 	37.98 & 	35.16 	\\
\textbf{Naive-RL} & 0.00 & 	0.00 	& 0.00 & 	0.00 & 	0.00 & 	0.00 &  96.01 & 	0.00 & 	0.00 & 	0.00 	& 10.87 	& 21.38 & 	10.69  \\
\textbf{SLC-RL}  & 76.00 	& 65.02 	& 78.94 & 	71.72 & 	80.71 	& 74.48 &  87.99 & 	67.70 	& 61.75 	& 69.34 & 	68.20 	& 71.00 	& 72.74 \\
\textbf{M-Thinker-4B $\Rightarrow$ Iter-1 (Ours)}	 &	\textbf{99.61} & 	\textbf{99.86} & 	\textbf{99.46} & 	\textbf{99.44} 	& \textbf{99.90} &  \textbf{99.66} &  99.42 	& 98.53 & 	\textbf{99.61} & 	99.33 & 	82.52 & 	95.88 	& \textbf{97.77} 	\\

\toprule
\multicolumn{14}{c}{\textbf{\textit{Metric:} Accuracy (Acc, \%)}} \\
\bottomrule
\rowcolor{gray!20}
\textbf{Qwen3-4B-Thinking-2507}  &		\textbf{76.15} & 	\textbf{77.30} & 	\textbf{78.70} 	& 79.17 & 	\textbf{76.06} 	& \textbf{77.48} &  84.36 & 	80.15 & 	74.51 & 	77.81 & 	75.42 & 	\textbf{78.45} & 	\textbf{77.96} \\

\textbf{Prompt-Control (No Training)} &	 75.72 	& 76.85 	& 78.10 & 	79.70 	& 75.37 & 	77.15&  \textbf{85.10} 	& 77.95 & 	73.66 & 	77.35 & 	74.30 & 	77.67 & 	77.41 	\\
\textbf{DIT (No Training)} &  70.63 & 	72.07 & 	77.88 & 	77.20 & 	75.44 	& 74.64  & 80.55 	& 78.72 	& 70.74 	& 73.01 & 	78.20 	& 76.24 & 	75.44 \\
\textbf{QRT (No Training)} &  69.16 & 	67.14 & 	75.78 &  	78.14 	& 66.80 	& 71.40  & 78.12 	& \textbf{80.65} 	& 66.30 & 	74.80 & 	76.86 & 	75.34 	& 73.37 \\

\textbf{Cold-Start SFT} &	64.41 & 	68.72 & 	78.55 & 	\textbf{80.24} & 	75.77 & 	73.54 &   83.05 & 	77.95 	& 76.09 	& 77.23 	& 76.90 & 	78.24 	& 75.89 	\\
\textbf{Naive-RL} & 76.07 	& 77.13 	& 77.57 & 	78.29 & 	75.82 & 	76.98 &  77.52 	& 78.11 & 	\textbf{76.32} & 	\textbf{79.03} & 	77.60 & 	77.72 & 	77.35 \\
\textbf{SLC-RL} & 58.78 	& 58.47 & 	71.82 & 	75.60 & 	60.89 & 	65.11 &  81.11 & 	75.37 & 	52.83 & 	68.25 	& 75.51 	& 70.61 & 	67.86 \\
\textbf{M-Thinker-4B $\Rightarrow$ Iter-1 (Ours)}	 &	71.41 & 	68.11 & 	76.38 & 	74.64 & 	67.23 & 	71.56 &  82.24 & 	78.63 	& 71.23 	& 73.27 & 	\textbf{78.57} & 	76.79 & 	74.17 	\\

\toprule
\multicolumn{14}{c}{\textbf{\textit{Metric:} Language Consistency \& Accuracy (LC\&Acc, \%)}} \\
\bottomrule
\rowcolor{gray!20}
\textbf{Qwen3-4B-Thinking-2507}  &	0.00 & 	0.00 & 	0.00 	& 0.00 &  	0.00 & 	0.00  & 84.36 & 	0.00 	& 0.00 	& 0.00 	& 47.40 & 	26.35 	& 13.18 	\\

\textbf{Prompt-Control (No Training)} &	 0.00 	& 0.00 	& 0.00 	& 0.00 	& 0.00 & 	0.00  & \textbf{85.03} & 	0.00 & 	0.02 & 	0.02 &  	49.59 & 	26.93 & 	13.47 	\\
\textbf{DIT (No Training)} &  66.67 & 	10.31 & 	2.37 & 	72.97 & 	8.84 & 	32.23 &  80.51 	& 76.86 & 	44.89 	& 72.35 & 	63.33 & 	67.59 & 	49.91 \\
\textbf{QRT (No Training)} &  68.96 	& 65.32 & 	73.68 & 	71.99 & 	64.08&  	68.81 &  78.10 & 	80.39&  	65.73 & 	\textbf{74.59} & 	63.91 & 	72.54 & 	70.68 \\

\textbf{Cold-Start SFT} &	 50.88 & 	41.79 & 	12.94&  	3.20 & 	13.12 	& 24.39  & 82.99 & 	8.50 & 	1.62 & 	17.40 	& 50.51 & 	32.21 & 	28.30 	\\
\textbf{Naive-RL} & 0.00 & 	0.00 	& 0.00 & 	0.00 & 	0.00 	& 0.00 &  74.20 & 	0.00 	& 0.00 	& 0.00 	& 10.46 & 	16.93 	& 8.47 \\
\textbf{SLC-RL}  & 56.25 & 	48.44 	& 66.37 & 	62.33 & 	59.68 & 	58.61 &   81.07 & 	59.93 & 	36.46 & 	51.86 & 	52.14 & 	56.29 & 	57.45 \\
\textbf{M-Thinker-4B $\Rightarrow$ Iter-1 (Ours)}	 &	\textbf{71.03} & 	\textbf{67.96} & 	\textbf{75.84} & 	\textbf{74.11} & 	\textbf{67.17}&  	\textbf{71.22} &  81.87 	& \textbf{77.51} 	& \textbf{70.85} & 	73.23 & 	\textbf{65.36} & 	\textbf{73.76} & 	\textbf{72.49} 	\\

\toprule

    \end{tabular}
    }
    \caption{
        The LC, Acc, and LC\&Acc (\%) results on the MMATH benchmark of the Qwen3-4B-Thinking-2507 backbone. ``\textit{\textbf{ID-avg}}/\textit{\textbf{OOD-avg}}'' is the average result of five In-Domain/Out-of-Domain languages and ``\textit{\textbf{ALL-AVG}}'' is the average result of all ten languages. The result in \textbf{bold} means the best result.
    }
    \label{table:main-res-4B}
\end{table*}

\begin{table*}[t]
    \centering
    \resizebox*{\linewidth}{!}{
    \begin{tabular}{l|ccccc|c|ccccc|c|c}
    \bottomrule
& \multicolumn{6}{c|}{\textbf{In-Domain Languages}} & \multicolumn{6}{c|}{\textbf{Out-of-Domain Languages}} & \\
\hline
\textbf{Methods} &	\textbf{ja}	&	\textbf{ko}	&	\textbf{fr}	&	\textbf{pt}	&	\textbf{th}	&	\textbf{\textit{ID-AVG}}	&	\textbf{en}	&	\textbf{es}	&	\textbf{ar}	&	\textbf{vi}	&	\textbf{zh}	&	\textbf{\textit{OOD-AVG}}	&	\textbf{\textit{ALL-AVG}}	\\

\toprule
\multicolumn{14}{c}{\textbf{\textit{Metric:} Language Consistency (LC, \%)}} \\
\bottomrule
\rowcolor{gray!20}

\textbf{DeepSeek-R1-Distill-Qwen-1.5B}  &	0.70 	 &	0.25 	 &	10.90  &		17.48 &	 	0.54 	&	5.98  & 91.01  &		17.68 	& 0.62  &		8.24 	 &	63.00 	 &			36.11  &		21.04 	\\

\textbf{Prompt-Control (No Training)} 	 & 4.41  & 	0.04  & 	20.35  & 	35.90  & 	2.49  & 	12.64  & 92.63  & 	40.93  & 	3.97  & 	39.89  & 	65.19  & 	48.52  & 	30.58 	\\


\textbf{DIT (No Training)} & 15.34 & 	0.29 & 	48.41 & 	44.85 & 	3.54 & 	22.48  & 90.25 & 	32.91 & 	4.18 & 	27.64 	& 61.50 	& 43.30 & 	32.89  \\
\textbf{QRT (No Training)} & 12.21 & 	0.08 & 	52.72 	& 41.60 & 	10.71 	& 23.47  & 90.97 & 	33.39 & 	9.34 & 	26.41 & 	67.19 & 	45.46 	& 34.46  \\

\textbf{Cold-Start SFT} &	1.81 &	0.00 &	49.82 &	54.34 	&12.68 &	23.73 &  90.39 &	42.53 &	2.01 &	26.06 &	77.77 &	47.75 &	35.74  \\
\textbf{Naive-RL}  &	0.00 &	0.00 &	0.00 &	0.00 &	0.00 &	0.00 & \underline{99.61} &	0.00 &	0.00 &	0.00 	& 55.23 &	30.97 &	15.48 \\
\textbf{SLC-RL}  & 0.00  &	0.00  &	0.00  &	0.00  &	0.00  &	0.00 & \textbf{100.00}  &	0.00  &	0.00  &	0.00  &	85.79  &	37.16  &	18.58  \\
\textbf{M-Thinker-1.5B $\Rightarrow$ Iter-1 (Ours)}	 &	\underline{98.68} &	\underline{98.17} &	\underline{99.54} &	\underline{99.70} &	\underline{99.84} &	\underline{99.19} & {98.44} &	\textbf{99.38} &	\textbf{33.31} &	\textbf{99.40} &	\underline{91.88} &	\textbf{84.48} &	\textbf{91.83}  \\
\textbf{M-Thinker-1.5B $\Rightarrow$ Iter-2 (Ours)}		& \textbf{99.76} &	\textbf{98.23} &	\textbf{99.73} 	& \textbf{99.84} 	& \textbf{99.88} &	\textbf{99.49} & {96.31} &	\underline{98.30} &	\underline{11.03} &	\underline{99.06} &	\textbf{92.86} &	\underline{79.51} &	\underline{89.50} 	\\

\toprule
\multicolumn{14}{c}{\textbf{\textit{Metric:} Accuracy (Acc, \%)}} \\
\bottomrule
\rowcolor{gray!20}
\textbf{DeepSeek-R1-Distill-Qwen-1.5B}  &34.28  &		32.48  &		36.91  &		39.22  &		31.17 	&	34.81  &	47.47  &		40.37  &		37.07  &		36.45  &		37.77  &				39.83  &		37.32  	\\

\textbf{Prompt-Control (No Training)} &	 	30.15  &	31.34  &	39.81  &	32.74  &	25.71  &	31.95  & 47.31  &	32.83  &	29.26  &	20.24  &	38.11  &	33.55  &	32.75 \\

\textbf{DIT (No Training)} & 19.39 &		17.41 &		31.51 &		28.66 &		18.53 &		23.10 &	 47.52 	&	27.10 &		11.49 &		17.31 &		39.09 &		28.50 &		25.80  \\
\textbf{QRT (No Training)} & 14.89 &		16.51 &		28.16 &		30.06 &		6.68 &		19.26 &	 45.55 &		26.10 	&	10.25 &		16.67 	&	42.01 	&	28.11 	&	23.69  \\

\textbf{Cold-Start SFT} &		  24.59 &	16.45 &	24.42 &	20.60 &	9.86 & 19.18 & 46.29 &	23.48 &	16.67 &	12.78 &	39.74 &	27.79 &	23.49 	\\
\textbf{Naive-RL} &	\textbf{51.12} & \textbf{50.15} &	\textbf{54.52} &	\textbf{52.58} &	\textbf{41.58} &	\textbf{49.99} &   {55.36} &	\textbf{53.83} &	\textbf{45.09} &	\textbf{47.70} &	48.45 &	\textbf{50.08} &	\textbf{50.04} 	\\
\textbf{SLC-RL}  &	\underline{46.69}  &		\underline{43.80}  &		\underline{54.23}  &		49.69 	 &	\underline{39.57} 	 &	\underline{46.80}  &	 \underline{56.37}  &		\underline{53.51} &	 	\underline{42.95} 	 &	\underline{46.11}  &		46.86  &		\underline{49.16} &	\underline{47.98}  \\

\textbf{M-Thinker-1.5B $\Rightarrow$ Iter-1 (Ours)}	&	 34.37 &	24.90 &	43.76 &	46.02 &	28.88 &	35.59 & 54.97 &	49.37 &	31.33 &	36.26 &	\underline{49.15} &	44.22 &	39.90  	\\
\textbf{M-Thinker-1.5B $\Rightarrow$ Iter-2 (Ours)}	&	{45.72} &	{33.40} 	& {50.02} &	\underline{51.63} &	{32.80} &	{42.72} & \textbf{56.51} &	{49.42} &	{37.14} 	& {37.73} &	\textbf{51.85} &	{46.53} &	{44.62}  	\\

\toprule
\multicolumn{14}{c}{\textbf{\textit{Metric:} Language Consistency \& Accuracy (LC\&Acc, \%)}} \\
\bottomrule
\rowcolor{gray!20}
\textbf{DeepSeek-R1-Distill-Qwen-1.5B}  &	0.22  &	0.02  &		7.05  &		11.92  &		0.12 &  3.87  &	46.56  &		13.38  &		0.16  &		3.56  &		32.30  &			19.19  &		11.53 	\\

\textbf{Prompt-Control (No Training)} &	 0.98  &		0.02 	 &	9.69  &		17.34  &		0.22  &		5.65  &	 46.42  &		19.65  &		0.62  &		13.52  &		31.40  &		22.32  &		13.99 	\\

\textbf{DIT (No Training)} & 7.83 &		0.06 &		25.99 &		23.36 &		0.55 &		11.56 &	 47.10 &		22.38 &		1.93 &		13.74 &		33.23 	&	23.68 &		17.62  \\
\textbf{QRT (No Training)} &  6.10 &		0.06 &		25.22 &		25.17 &		1.86 &		11.68 &	 45.45 	&	21.61 &		2.48 &		13.66 &		36.17 &		23.87 	&	17.78 \\

\textbf{Cold-Start SFT} &	1.11 &	0.00 &	17.29 &	16.99 &	1.56  &  7.39 & 45.84 	& 20.54 &	0.52 &	7.25 	& 34.51  & 21.73 &	14.56 	\\
\textbf{Naive-RL} &	 0.00  &		0.00  &		0.00  &		0.00  &		0.00 	 &	0.00  &	 {55.31}  &		0.00  &		0.00  &		0.00  &		25.47  &		16.16  &		8.08 	\\
\textbf{SLC-RL} &	 0.00 &		0.00 &		0.00 	&	0.00 &		0.00 & 0.00 & 		\underline{56.37} &		0.00 	&	0.00 &		0.00 &		40.99 &		19.47 &		9.74  \\

\textbf{M-Thinker-1.5B $\Rightarrow$ Iter-1 (Ours)}	 &	\underline{34.25}  &		\underline{24.48}  &		\underline{43.72}  &		\underline{45.78}  & 	\underline{28.72}  & 	\underline{35.39}  &  54.89 	 & \underline{49.19}  & 	\textbf{6.39}  & 	\underline{35.76}  & 	\underline{45.60}  & 	\underline{38.37}  & 	\underline{36.88} 	\\
\textbf{M-Thinker-1.5B $\Rightarrow$ Iter-2 (Ours)}	&	\textbf{45.54} &	\textbf{32.86} &	\textbf{49.75} &	\textbf{51.47} &	\textbf{32.72} &	\textbf{42.47} & \textbf{56.41} &	\textbf{49.20} &	\underline{2.80} &	\textbf{37.55} &	\textbf{48.20} &	\textbf{38.83} &	\textbf{40.65}  	\\

\toprule

    \end{tabular}
    }
    \caption{
        The LC, Acc, and LC\&Acc (\%) results on the MMATH benchmark of the DeepSeek-R1-Distill-Qwen-1.5B backbone. ``\textit{\textbf{ID-avg}}/\textit{\textbf{OOD-avg}}'' is the average result of five In-Domain/Out-of-Domain languages and ``\textit{\textbf{ALL-AVG}}'' is the average result of all ten languages. The result in \textbf{bold} means the best result, and the \underline{underlined} result means the second-best result in each setting.
        ``\textbf{Iter-1/2}'' means the training iteration 1/2.
    }
    \label{table:main-res-1.5B}
\end{table*}

\begin{table*}[t]
    \centering
    \resizebox*{\linewidth}{!}{
    \begin{tabular}{l|ccc|ccc|ccc}
    \bottomrule
     & \multicolumn{3}{c|}{\textbf{LC}} & \multicolumn{3}{c|}{\textbf{Acc}} & \multicolumn{3}{c}{\textbf{LC\&Acc}}\\
    \hline
    \textbf{Settings} & \textbf{\textit{ID-AVG}} & \textbf{\textit{OOD-AVG}} & \textbf{\textit{ALL-AVG}} & \textbf{\textit{ID-AVG}} & \textbf{\textit{OOD-AVG}} & \textbf{\textit{ALL-AVG}} & \textbf{\textit{ID-AVG}} & \textbf{\textit{OOD-AVG}} & \textbf{\textit{ALL-AVG}} \\
    \hline

\textit{w/} Hard-LC (1.5B) & \textbf{99.16} & \textbf{92.44} & \textbf{95.80} & 31.72 & 39.85 & 35.78 & \textbf{31.68} & \textbf{37.18} & \textbf{34.43} \\
\textit{w/} Soft-LC (1.5B) &	0.00	& 37.16	& 18.58	& \textbf{46.80}	& \textbf{49.16}	& \textbf{47.98}	& 0.00	& 19.47	& 9.74 \\
\hline
\textit{w/} Hard-LC (7B)	& \textbf{99.34}	& \textbf{92.27}	& \textbf{95.80}	& 55.56	& \textbf{62.09}	& \textbf{58.83}	& \textbf{55.47}	& \textbf{58.16} & \textbf{56.81} \\
\textit{w/} Soft-LC (7B)	& 76.00	& 74.24	& 75.12	& \textbf{56.64}	& 60.55	& 58.59	& 43.14	& 46.88	& 45.01 \\

    \toprule
    \end{tabular}
    }
    \caption{
        The detailed comparison between the "Hard-LC" ( with LC/Format/Acc rewards) and "Soft-LC" (with SLC/Format/Acc rewards).
    }
    \label{table:hard-soft-results}
\end{table*}

\begin{table*}[t]
    \centering
    \resizebox*{\linewidth}{!}{
    \begin{tabular}{ccc|cccc}
    \bottomrule

\textbf{Format} &	\textbf{Language Consistency}	&	\textbf{Accuracy}   & \textbf{SLC-RL} & \textbf{SLC-RL-s} & \textbf{Ours} (\textit{w/o} $R_{\text{cta}}$)  & \textbf{Ours} \\
\hline
\ding{55} & / & / & 0 & -1 & -1 & -1 \\
$\checkmark$ & \ding{55} & \ding{55} & 1*(0+0)=0 & 1*(-1+0)=-1 & -1 & -1 \\
$\checkmark$ & \ding{55} & $\checkmark$ & 1*(0+0.9)=0.9 & 1*(-1+1)=0 & -1 & -1 \\
$\checkmark$ & $\checkmark$ & \ding{55} & 1*(0.1+0)=0.1 & 1*(0+0)=0 & 0*1=0 & 0*(1+$R_{\text{cta}}$)=0 \\
$\checkmark$ & $\checkmark$ & $\checkmark$ & 1*(0.1+0.9)=1 & 1*(0+1)=1 & 1*1=1 & 1*(1+$R_{\text{cta}}$)=1+$R_{\text{cta}}$ \\

\toprule

    \end{tabular}
    }
    \caption{
        The detailed reward scores of different methods on different scenarios. ``SLC-RL-s'' means the same scale of soft-LC reward with our $R_{\text{lc}}$. ``Ours'' represents M-thinker-1.5B (Iter-1).
    }
    \label{table:dif-reward-scores}
\end{table*}

\begin{table*}[h]
    \centering
    \resizebox*{\linewidth}{!}{
    \begin{tabular}{l|c|ccc|ccc|ccc}
    \bottomrule
     &  & \multicolumn{3}{c|}{\textbf{LC}} & \multicolumn{3}{c|}{\textbf{Acc}} & \multicolumn{3}{c}{\textbf{LC\&Acc}}\\
    \hline
    \textbf{Methods} & \textbf{LC/Acc Reward} & \textbf{\textit{ID-AVG}} & \textbf{\textit{OOD-AVG}} & \textbf{\textit{ALL-AVG}} & \textbf{\textit{ID-AVG}} & \textbf{\textit{OOD-AVG}} & \textbf{\textit{ALL-AVG}} & \textbf{\textit{ID-AVG}} & \textbf{\textit{OOD-AVG}} & \textbf{\textit{ALL-AVG}} \\
    \hline
SLC-RL & [0,0.1]+[0,0.9] & 0.00 &	37.16	& 18.58	& 46.80 &	49.16 &	47.98 &	0.00 &	19.47 &	9.74 \\
SLC-RL-s & [-1,0]+[0,1] &	39.34 &	57.46 &	48.40 &	44.66 &	48.78 &	46.72 &	16.95 &	28.96 &	22.96 \\
Ours (\textit{w/o} $R_{\text{cta}}$) & [-1,0]\&[0,1] &	99.16 &	92.44 &	95.80 &	31.72 &	39.85 &	35.78 &	31.68 &	37.18 &	34.43 \\
Ours & [-1,0]\&[0,1](+$R_{\text{cta}}$) &  99.19 & 84.48 & 91.83 &  35.59 & 44.22 & 39.90 & 35.39 & 38.37 & 36.88 \\

    \toprule
    \end{tabular}
    }
    \caption{
        The results of SLC-RL with different reward magnitudes. ``SLC-RL-s'' means the same scale of soft-LC reward with our $R_{\text{lc}}$. ``Ours'' represents M-thinker-1.5B (Iter-1).
    }
    \label{table:slc-results}
\end{table*}

\begin{table*}[t]
    \centering
    \resizebox*{\linewidth}{!}{
    \begin{tabular}{l|ccccc|c|ccccc|c|c}
    \bottomrule
& \multicolumn{6}{c|}{\textbf{In-Domain Languages}} & \multicolumn{6}{c|}{\textbf{Out-of-Domain Languages}} & \\
\hline
\textbf{Methods} &	\textbf{ja}	&	\textbf{ko}	&	\textbf{fr}	&	\textbf{pt}	&	\textbf{th}	&	\textbf{\textit{ID-AVG}}	&	\textbf{en}	&	\textbf{es}	&	\textbf{ar}	&	\textbf{vi}	&	\textbf{zh}	&	\textbf{\textit{OOD-AVG}}	&	\textbf{\textit{ALL-AVG}}	\\

\toprule
\multicolumn{14}{c}{\textbf{\textit{Metric:} Language Consistency (LC, \%)}} \\
\bottomrule
\rowcolor{gray!20}
\textbf{DeepSeek-R1-Distill-Qwen-1.5B} &  7.30  & 	0.15  & 	25.65  & 	25.80 	 & 8.45  & 	13.47   & 91.30 	 & 26.55  & 	9.05 	 & 22.90  & 	63.35  & 	42.63  & 	28.05 \\
\textbf{M-Thinker-1.5B $\Rightarrow$ Iter-1 (Ours)} &  98.25  & 	96.40  & 	\textbf{99.85}  & 	99.00 	 & 99.70  & 	98.64  &  97.40  & 	\textbf{99.40}  & 	\textbf{40.40}  & 	97.50  & 	88.60  & 	\textbf{84.66} 	 & \textbf{91.65}  \\
\textbf{M-Thinker-1.5B $\Rightarrow$ Iter-2 (Ours)}	 & \textbf{99.40}  & 	\textbf{98.65}  & 	99.80  & 	99.00  & 	\textbf{99.85}  & 	\textbf{99.34}  &  \textbf{97.50}  & 	98.90 	 & 19.65  & 	\textbf{99.25}  & 	\textbf{90.10}  & 	81.08  & 	90.21 	\\
\hline
\rowcolor{gray!20}
\textbf{DeepSeek-R1-Distill-Qwen-7B} & 20.85  & 	11.35  & 	26.80  & 	24.10  & 	14.85  & 	19.59  &  96.05  & 	26.20  & 	14.90 	 & 26.30 	 & 67.70  & 	46.23  & 	32.91  \\
\textbf{M-Thinker-7B $\Rightarrow$ Iter-1 (Ours)} &  \textbf{99.05}  & 	\textbf{97.65}  & 	\textbf{{99.85}}  & 	\textbf{99.25}  & 	\textbf{98.40}  & 	\textbf{98.84}  &  \textbf{99.80} 	 & \textbf{99.65}  & 	\textbf{83.75}  & 	{99.80}  & 	\textbf{89.70}  & 	\textbf{94.54}  & 	\textbf{96.69}  \\
\textbf{M-Thinker-7B $\Rightarrow$ Iter-2 (Ours)} & 98.75 	& 95.30 	& 99.65 & 	99.00 & 	94.65 &  97.47 & 97.55 & 	98.65 & 	64.80 & 	\textbf{100.00} & 	89.25 & 	90.05 & 	93.76  \\

\toprule
\multicolumn{14}{c}{\textbf{\textit{Metric:} Difficulty-Weighted Accuracy (DW-ACC, \%)}} \\
\bottomrule
\rowcolor{gray!20}
\textbf{DeepSeek-R1-Distill-Qwen-1.5B} & 13.60  & 	15.77  & 	18.62  & 	18.73  & 	11.25 	 & 15.59  &  21.23  & 	19.47  & 	14.36  & 	18.25  & 	20.00  & 	18.66  & 	17.13 \\
\textbf{M-Thinker-1.5B $\Rightarrow$ Iter-1 (Ours)} &  16.64  & 	12.33  & 	23.03  & 	23.40  & 	12.77  & 	17.63  &  30.13  & 	23.23  & 	15.90  & 	17.42  & 	25.08  & 	22.35 	 & 19.99  \\
\textbf{M-Thinker-1.5B $\Rightarrow$ Iter-2 (Ours)}	& \textbf{19.65}  & 	\textbf{17.39}  & 	\textbf{24.87}  & 	\textbf{24.76}  & 	\textbf{16.84}  & 	\textbf{20.70} &   \textbf{32.41}  & 	\textbf{25.63}  & 	\textbf{19.11}  & 	\textbf{20.83}  & 	\textbf{27.98}  & 	\textbf{25.19}  & 	\textbf{22.95} 	\\
\hline

\rowcolor{gray!20}
\textbf{DeepSeek-R1-Distill-Qwen-7B}  &   28.45  & 	31.72  & 	35.41  & 	33.12  & 	27.72 &  	31.28  &  36.93  & 	33.51  & 	28.93  & 	31.96  & 	30.67  & 	32.40 	 & 31.84 	\\

\textbf{M-Thinker-7B $\Rightarrow$ Iter-1 (Ours)} & 29.99  & 	28.59  & 	35.02  & 	35.30  & 	29.35  & 	31.65  &  40.80  & 	34.72  & 	29.83  & 	31.99  & 	34.80  & 	34.43  & 	33.04 \\
\textbf{M-Thinker-7B $\Rightarrow$ Iter-2 (Ours)} & \textbf{35.24} & 	\textbf{33.92} & 	\textbf{40.02} & 	\textbf{39.73} 	& \textbf{34.40} 	& \textbf{36.66}  & \textbf{42.48} & 	\textbf{38.33} & 	\textbf{37.08} & 	\textbf{37.19} 	& \textbf{42.73} & 	\textbf{39.56} & 	\textbf{38.11} \\

\toprule
\multicolumn{14}{c}{\textbf{\textit{Metric:} Language Consistency \& Difficulty-Weighted Accuracy (LC\&DW-ACC, \%)}} \\
\bottomrule
\rowcolor{gray!20}
\textbf{DeepSeek-R1-Distill-Qwen-1.5B} & 0.67  & 	0.00  & 	3.03  & 	3.36  & 	0.16  & 	1.44   & 21.15  & 	3.71  & 	0.57 &  	2.40 &  	16.09  & 	8.78  & 	5.11  \\
\textbf{M-Thinker-1.5B $\Rightarrow$ Iter-1 (Ours)} &  16.43 	 & 12.07 	 & 23.03  & 	23.27  & 	12.75  & 	17.51   & 29.89  & 	23.18  & 	\textbf{4.55}  & 	16.96  & 	22.87 	 & 19.49 	 & 18.50   \\
\textbf{M-Thinker-1.5B $\Rightarrow$ Iter-2 (Ours)} & 	\textbf{19.61}  & 	\textbf{17.04}  & 	\textbf{24.86} 	 & \textbf{24.51}  & 	\textbf{16.80}  & 	\textbf{20.56}  &   \textbf{32.36}  & 	\textbf{25.44}  & 	1.55  & 	\textbf{20.59}  & 	\textbf{24.62}  & 	\textbf{20.91}  & 	\textbf{20.74} 	\\
\hline
\rowcolor{gray!20}
\textbf{DeepSeek-R1-Distill-Qwen-7B}  &  3.05  & 	1.89  & 	5.89  & 	4.44  & 	2.39  & 	3.53   & 36.70  & 	5.74  & 	2.45  & 	4.89 	 & 25.27 	 & 15.01  & 	9.27 \\

\textbf{M-Thinker-7B $\Rightarrow$ Iter-1 (Ours)}	 &  29.81  & 	28.55  & 	34.97  & 	35.11  & 	29.11 &  	31.51  &  40.67  & 	34.61  & 	\textbf{27.18}  & 	31.97  & 	31.73  & 	33.23  & 	32.37  \\
\textbf{M-Thinker-7B $\Rightarrow$ Iter-2 (Ours)} &  \textbf{35.13} &  	\textbf{33.13} & 	\textbf{39.96} 	& \textbf{39.36} 	& \textbf{33.79} 	& \textbf{36.27}  & \textbf{42.48} 	& \textbf{37.96} 	& 24.18 	& \textbf{37.19} 	& \textbf{39.24} 	& \textbf{36.21} 	& \textbf{36.24} \\

\toprule

    \end{tabular}
    }
    \caption{
        The LC, DW-ACC, and LC\&DW-ACC (\%) results on the PolyMath benchmark of the DeepSeek-R1-Distill-Qwen-1.5B/7B backbones. The result in \textbf{bold} means the best result in each backbone.
    }
    \label{table:main-res-polymath}
\end{table*}

\subsection{Results of Qwen3-4B-Thinking-2507}\label{sec:appendix-qwen3}
We list the LC, Acc, and LC\&Acc (\%) results on the MMATH benchmark of the Qwen3-4B-Thinking-2507 backbone in Table \ref{table:main-res-4B}.
The results demonstrate the superiority of our method on both language consistency and answer accuracy over other baselines.
Due to the time and resource limitations, we only conduct GRPO training for one iteration.
The effectiveness of the iterative training strategy has been proven in the other backbones.

\subsection{Detailed Results of DeepSeek-R1-Distill-Qwen-1.5B}\label{sec:appendix-res-qwen1.5}
We report the LC, Acc, and LC\&Acc (\%) results on the MMATH benchmark of the DeepSeek-R1-Distill-Qwen-1.5B backbone for each language in Table \ref{table:main-res-1.5B}.

\subsection{Hard vs. Soft Language Consistency Reward}\label{sec:appendix-hard-lc}
To compare the performance of hard/soft LC reward, we conduct GRPO training from the same cold-start SFT model and report the results in Table \ref{table:hard-soft-results}.
The results reveal distinct behaviors across model sizes.
For the 1.5B Model, the hard LC constraint ensures higher language consistency but leads to a drop in answer accuracy (Acc) compared to the soft LC reward, as the smaller model struggles to satisfy strict language constraints while reasoning correctly.
For the 7B Model, the accuracy gap becomes negligible (the ALL-AVG ``58.83'' of Hard-LC even surpasses the one ``58.59'' of Soft-LC). 
Additionally, the Hard-LC reward significantly outperforms the Soft-LC reward in maintaining language consistency (``95.8'' vs. ``75.12'').
In summary, the performance achieved with either hard-LC or soft-LC ultimately depends on the model's inherent capability: the stronger the model, the better hard-LC can simultaneously ensure linguistic consistency and answer accuracy.

\subsection{SLC-RL with Different Reward Magnitudes}\label{sec:appendix-soft-lc}
For a clear comparison, we also conduct SLC-RL with the same reward magnitude as ours. 
First, we list the detailed reward scores of different methods in Table \ref{table:dif-reward-scores}.
Specifically, ``SLC-RL-s'' utilizes the same LC-reward magnitude as our method; however, it still underperforms our method as shown in Table \ref{table:slc-results}. 
The results demonstrate that the hard LC reward facilitates higher language consistency, and the effectiveness of our M-Thinker stems from the strict LC reward and the CTA reward rather than merely the LC reward scale.

\subsection{Results of PolyMath}\label{sec:appendix-polymath}
We report the LC, DW-ACC, and LC\&DW-ACC (\%) evaluation results on the PolyMath benchmark of the DeepSeek-R1-Distill-Qwen-1.5B/7B backbones in Table \ref{table:main-res-polymath}.
These results also demonstrate the superiority of our M-Thinker-1.5B/7B.

\section{Additional Analysis}

\subsection{Detailed Ablation Results}\label{sec:appendix-detailed-ablation-res}

We list the detailed ablation results of the MMATH benchmark based on our M-Thinker-1.5B (Iter-1) in Table \ref{table:ablation-detailed-res}.


\subsection{Other Alternative Judge Metrics of the CTA Reward}\label{sec:appendix-clarify-dif-metric}
MAPO \cite{she2024mapoadvancingmultilingualreasoning} utilizes the NLLB model to calculate translation probabilities between English and multilingual responses, serving as a selection criterion to construct preference data for DPO training.
While this translation-consistency approach can be adapted as a reward function for cross-lingual alignment, we suspect that it may not be an effective additional reward signal from two primary concerns:
(1) Translation models often degrade when processing long chain-of-thought (CoT) reasoning heavily interspersed with mathematical formulas (LaTeX). This may result in erroneous reward signals and lead to unstable RL training. As evidenced in Table 4 of CM-Align \cite{zhang2025cmalignconsistencybasedmultilingualalignment}, MAPO struggles to maintain high reward accuracy across different models.
(2) Fundamentally, a translation-based reward encourages the model to generate literal translations of the English reasoning path. This forces the model into "translationese" rather than allowing it to ``think natively'' in the target language. 
In contrast, our CTA reward checks for the logical equivalence of key intermediate steps. 
This allows the model to employ native reasoning patterns and syntactic structures as long as the key steps remain correct.


\begin{table*}[ht]
    \centering
    \resizebox*{\linewidth}{!}{
    \begin{tabular}{l|ccccc|c|ccccc|c|c}
    \bottomrule
& \multicolumn{6}{c|}{\textbf{In-Domain Languages}} & \multicolumn{6}{c|}{\textbf{Out-of-Domain Languages}} & \\
\hline
\textbf{Methods} &	\textbf{ja}	&	\textbf{ko}	&	\textbf{fr}	&	\textbf{pt}	&	\textbf{th}	&	\textbf{\textit{ID-AVG}}	&	\textbf{en}	&	\textbf{es}	&	\textbf{ar}	&	\textbf{vi}	&	\textbf{zh}	&	\textbf{\textit{OOD-AVG}}	&	\textbf{\textit{ALL-AVG}}	\\

\toprule
\multicolumn{14}{c}{\textbf{\textit{Metric:} Language Consistency (LC, \%)}} \\
\bottomrule
\rowcolor{gray!20}
\textbf{M-Thinker-1.5B $\Rightarrow$ Iter-1 (Ours)}	 &	{98.68} &	{98.17} &	{99.54} &	{99.70} &	{99.84} &	{99.19} & {98.44} &	{99.38} &	{33.31} &	{99.40} &	{91.88} &	{84.48} &	{91.83}  \\
\textbf{\quad \quad \textit{w/o} $R_{\text{cta}}$}  &	 99.40  &		97.59  &		99.16  &		99.67  &		99.98  &		99.16  &	 98.61  &		98.91  &		73.88  &		99.96  &		90.84  &		92.44  &		95.80  \\
\textbf{\quad \quad \textit{w/o} $R_{\text{lc}}$} &	0.00 &		0.00 &		0.00 &		0.00 &		0.00 &		0.00 &	 99.88 	&	0.00 &		0.00 &		0.00 &		78.17 &		35.61 &		17.80  \\
\textbf{\quad \quad \textit{w/o} ($R_{\text{cta}}$ \& $R_{\text{lc}}$)} &	0.00 &	0.00 &	0.00 &	0.00 &	0.00 &	0.00 & {99.61} &	0.00 &	0.00 &	0.00 	& 55.23 &	30.97 &	15.48 \\

\textbf{\quad \quad \textit{w/o} Cold-Start SFT} & 99.23  &		99.02  &		98.37  &		99.42  &		99.90  &		99.19  & 95.37  &		99.59  &		35.93  &		99.63  &		91.14  &		84.33  &		91.76 \\
\textbf{\quad \quad \textit{w/o} Rejection Sampling} & 99.24  &		99.68  &		99.98  &		99.80 	 &	99.86 	 &	99.71   &	 99.31  &		98.55 	 &	40.24  &		97.32  &		91.12  &		85.31 	 &	92.51    \\
\textbf{\quad \quad \textit{w/} $o_t^{en}$ from Light-R1 for $R_{\text{cta}}$}  &	99.98  &		99.82  &		99.52 	 &	99.46  &		100.00  &		99.76  &	 99.73  &		99.61  &		49.19  &		99.55 	 &	93.26  &		88.27  &		94.01  \\

\toprule
\multicolumn{14}{c}{\textbf{\textit{Metric:} Accuracy (Acc, \%)}} \\
\bottomrule
\rowcolor{gray!20}

\textbf{M-Thinker-1.5B $\Rightarrow$ Iter-1 (Ours)}	&	 34.37 &	24.90 &	43.76 &	46.02 &	28.88 &	35.59 & 54.97 &	49.37 &	31.33 &	36.26 &	{49.15} &	44.22 &	39.90  	\\
\textbf{\quad \quad \textit{w/o} $R_{\text{cta}}$}  &	30.48  &		23.75  &		39.45  &		41.34  &		23.59  &		31.72  &	 51.87  &		43.26  &		27.99  &		31.11  &		45.01  &		39.85 	 &	35.78  \\
\textbf{\quad \quad \textit{w/o} $R_{\text{lc}}$}  & 49.53  &	47.06  &	56.48  &	53.18  &	44.84  &	50.22  & 57.40  &	55.27  &	43.55  &	50.37  &	47.54  &	50.83  &	50.52 \\
\textbf{\quad \quad \textit{w/o} ($R_{\text{cta}}$ \& $R_{\text{lc}}$)}  & 51.12 	 & 50.15  & 	54.52  & 	52.58  & 	41.58 	 & 49.99  &  55.36  & 	53.83 	 & 45.09  & 	47.70  & 	48.45 &  	50.08  & 	50.04  \\

\textbf{\quad \quad \textit{w/o} Cold-Start SFT} & 31.18 	 & 22.15  & 	42.44 	 & 45.25  & 	26.99  & 	33.60  & 52.68  & 	45.66  & 	31.31 &  	34.15  & 	50.35  & 	42.83  & 	38.22 \\
\textbf{\quad \quad \textit{w/o} Rejection Sampling} & 34.40  & 	19.17  & 	45.75  & 	44.63  & 	25.41  & 	33.87 & 54.55  & 	43.41 	 & 28.16  & 	33.42 	 & 46.66  & 	41.24  & 	37.55  \\
\textbf{\quad \quad \textit{w/} $o_t^{en}$ from Light-R1 for $R_{\text{cta}}$}  &	31.37 	 &	24.88 	 &	42.02  &		43.18  &		27.10  &		33.71 &	 54.43  &		46.96  &		28.53  &		33.06  &		46.37  &		41.87 	 &	37.79   \\

\toprule
\multicolumn{14}{c}{\textbf{\textit{Metric:} Language Consistency \& Accuracy (LC\&Acc, \%)}} \\
\bottomrule
\rowcolor{gray!20}

\textbf{M-Thinker-1.5B $\Rightarrow$ Iter-1 (Ours)}	 &	{34.25}  &		{24.48}  &		{43.72}  &		{45.78}  & 	{28.72}  & 	{35.39}  &  54.89 	 & {49.19}  & 	{6.39}  & 	{35.76}  & 	{45.60}  & 	{38.37}  & 	{36.88} 	\\
\textbf{\quad \quad \textit{w/o} $R_{\text{cta}}$}  &	30.46  &		23.73  &		39.41  &		41.24  &		23.57  &		31.68  &	 51.77 &	 	42.67  &		19.76  &		31.09  &		40.63  &		37.18  &		34.43  \\
\textbf{\quad \quad \textit{w/o} $R_{\text{lc}}$}  & 0.00  &	0.00  &	0.00  &	0.00  &	0.00  &	0.00  & 57.28 	 & 0.00  &	0.00  &	0.00  &	36.03  &	18.66  &	9.33 \\
\textbf{\quad \quad \textit{w/o} ($R_{\text{cta}}$ \& $R_{\text{lc}}$)}  & 0.00  & 	0.00  & 	0.00  & 	0.00  & 	0.00  & 	0.00  &  55.31 &  	0.00  & 	0.00  & 	0.00 	 & 25.47 	 & 16.16  & 	8.08 \\
\textbf{\quad \quad \textit{w/o} Cold-Start SFT} & 31.00  & 	21.77 	 & 42.19  & 	44.88  & 	26.89  & 	33.35  & 50.60  & 	45.50  & 	8.93 	 & 34.09  & 	45.43 	 & 36.91  & 	35.13 \\
\textbf{\quad \quad \textit{w/o} Rejection Sampling} & 34.20  & 	19.01  & 	45.73  & 	44.45  & 	25.27  & 	33.73  &  53.86  & 	42.37  & 	5.81  & 	32.65  & 	42.68 	 & 35.48  & 	34.60  \\
\textbf{\quad \quad \textit{w/} $o_t^{en}$ from Light-R1 for $R_{\text{cta}}$}  &	31.35 	 &	24.80  &		42.00  &		43.10  &		27.10  &		33.67  &	 54.39  &		46.63  &		10.90  &		32.63  &		43.68  &		37.65  &		35.66  \\

\toprule

    \end{tabular}
    }
    \caption{
        The detailed ablation results of the MMATH benchmark based on our M-Thinker-1.5B (Iter-1). 
    }
    \label{table:ablation-detailed-res}
\end{table*}

\subsection{Annotation Guidelines of Human Evaluation}\label{sec:appendix-human}
The annotation follows a rigorous two-step procedure:
\begin{itemize}
    \item \textbf{Reference Extraction:} First, experts proficient in English extract key intermediate reasoning steps from the reference English thinking paths.
    \item \textbf{Alignment Verification:} Experts proficient in the target languages are then provided with the non-English thinking paths and the extracted English key steps. They independently identify key intermediate steps in the target language and calculate the alignment ratio with the English counterparts.
\end{itemize}

\end{document}